# Analysis of the Efficacy of the Use of Inertial Measurement and Global Positioning System Data to Reverse Engineer Automotive CAN Bus Steering Signals

Kevin Setterstrom
Department of Computer Science
North Dakota State University
1320 Albrecht Blvd., Room 258
Fargo, ND 58108

P: +1 (701) 231-8562
F: +1 (701) 231-8255
E: kevin.setterstrom@ndsu.edu

Jeremy Straub
Department of Computer Science
North Dakota State University
1320 Albrecht Blvd., Room 258
Fargo, ND 58108

P: +1 (701) 231-8196
F: +1 (701) 231-8255
E: jeremy.straub@ndsu.edu

**Abstract**

Autonomous vehicle control is growing in availability for new vehicles and there is a potential need to retrofit older vehicles with this capability. Additionally, automotive cybersecurity has become a significant concern in recent years due to documented attacks on vehicles. As a result, researchers have been exploring reverse engineering techniques to automate vehicle control and improve vehicle security and threat analysis. In prior work, a vehicle's accelerator and brake pedal controller area network (CAN) channels were identified using reverse engineering techniques without prior knowledge of the vehicle. However, the correlation results for deceleration were lower than those for acceleration, which may be able to be improved by incorporating data from an additional telemetry device. In this paper, a method that uses IMU and GPS data to reverse-engineer a vehicle's steering wheel position CAN channels, without prior knowledge of the vehicle, is presented. Using GPS data is shown to greatly improve correlation values for deceleration, particularly for the brake pedal CAN channels. This work demonstrates the efficacy of using these data sources for automotive CAN reverse engineering. This has potential uses in automotive vehicle control and for improving vehicle security and threat analysis.

## 1. Introduction

Automotive controller area network (CAN) reverse engineering has been a topic of growing interest during the last few years. Reverse engineering can be used to enable vehicle control. Several examples of vehicles being attacked have been documented [1]–[7], creating interest from a security perspective. These attacks have illustrated a number of new threat vectors and presented new research opportunities. In prior work [8], it was shown that the CAN channels pertaining to a vehicle's accelerator and brake pedals could be autonomously reverse engineered, without requiring prior knowledge of the vehicle. The system proposed previously correlated IMU data and CAN channels and used reverse engineering techniques to identify the specific channels related to pedal controls. However, the correlation results for the vehicle's deceleration were found to be lower than those for acceleration. The correlation value is important to this method. Therefore, new methods that improve correlation results would likely improve the system's performance. This is especially true with regards to the scalability and applicability of the system to all automotive manufacturers.

This paper assesses whether incorporating data from an additional telemetry device will enhance the identification of vehicle actions. It, thus, incorporates GPS data along with the IMU data that was used previously. The aim of this is to improve on previous work by including GPS data in the preprocessing phase of the algorithms and provide improved results for deceleration correlation performance. In the previous work [8], the reverse engineering of steering wheel CAN channels was not examined. This paper also builds on the previous research by using IMU and GPS data to reverse engineer the vehicle's steering wheel position CAN channels.

Being able to monitor, understand, and set a vehicle's steering wheel position can be useful for developing an aftermarket self-driving vehicle conversion kit and for performing vehicle threat analysis. However, the specific CAN information required to do this is not publicly available, so the ability to quickly and easily reverse engineer these CAN messages is advantageous for developing control systems and systems to monitor and improve vehicle security. The work presented in this paper can be used for identifying vehicle CAN messages related to vehicle controls without requiring prior information about the vehicle. It builds upon the work presented in [8] by increasing deceleration correlation accuracy and introducing a capability for steering wheel position message determination.

## 2. Background

This section provides background information regarding the areas built upon by this work. It begins by discussing the importance of automotive CAN reverse engineering in the modern automotive industry. It has a focus on the growing integration of drive-by-wire systems and the vulnerabilities that arise in connected vehicles. The section then highlights significant research endeavors in the field, including prior studies on vulnerabilities in vehicle CAN systems and different approaches to automating the reverse engineering process. Finally, the section concludes with a brief summary of the previous work [8] that this research directly builds upon and enhances.

### 2.1. Importance of CAN Reverse Engineering

The development of drive-by-wire technology, which uses CAN bus messages as part of a vehicle's control system, has led to advancements in automotive vehicle controls, drivability, and safety. Isermann, et al. [9] discussed the importance of reliability and safety in drive-by-wire systems. They noted that engineering challenges include designing mass-producible fault-tolerant sensors, actuators, microcomputers, and bus communication systems at a reasonable cost. Throttle-by-wire, brake-by-wire, and steer-by-wire are commonly discussed automotive advancements that are part of drive-by-wire technology. Throttle-by-wire is already widely used, while brake-by-wire and steer-by-wire have not yet been as widely implemented. Due to their criticality, backup systems are used to protect these systems against faults; however, this increases complexity and development cost. The main impediment to the widespread adoption of steer-by-wire is concern regarding reliability and public mistrust. Electronics' reliability and functional safety need to be improved before standalone steer-by-wire can be introduced [9]–[11].

Drive-by-wire technology has facilitated the addition of advanced driver assistance systems (ADAS), which have improved the safety of automotive transportation through sensor-based and connected vehicle systems. Features such as autonomous emergency braking (AEB), blind spot warning/lane change warning (BSW/LCW), forward collision warning (FCW), and lane departure warning (LDW) communicate their data using the vehicle's CAN bus and provide important autonomous safety features in vehicles [12], [13].

Drive-by-wire technology requires CAN messages identifying the position of vehicle inputs such as the accelerator pedal, brake pedal, and steering wheel. If these messages can be reverse engineered, aftermarket vehicle control systems can be developed. If they are tampered with, unwanted outcomes, especially when moving at high speeds, could occur. The potential issues and benefits show the importance and potential impact that automotive CAN bus reverse engineering could have.

### 2.2. Prior Work in CAN Reverse Engineering

In the automotive industry, the introduction of self-driving and highly connected vehicles has created new security challenges [14]. Malicious actors can exploit vulnerabilities in vehicle CAN systems, as demonstrated by researchers like Miller and Valasek [1]–[3], who were able to gain control of a 2014 Jeep Cherokee remotely [3]. Their study was significant because it countered the widespread belief that such attacks required close physical access to the vehicle. This research contributed to the broader understanding and discussion regarding the security of connected vehicles. It highlights the importance of considering remote vulnerabilities in automotive security frameworks.

In a multi-purpose effort to address these issues, as well as providing a means for aftermarket vehicle autonomy, researchers have proposed automated reverse engineering methods for CAN data. The proposed methods include Marchetti and Stabili's READ algorithm [14], which automatically identifies signal boundaries in CAN traffic, and Huybrechts, et al.'s [15] solution that compares known OBD-II PID signals with raw CAN data. Kang, et al. [16] introduced the automated CAN analyzer (ACA), which identifies and injects identified messages into the CAN bus. Pese, et al. developed LibreCAN [12], which uses DBC files, OBD-II PID messages, and cross correlation to successfully decode important CAN information. In another effort, Blaauwedraad and Kieberl [17] developed an automated reverse-engineering method which utilizes the Pearson correlation coefficient to find a direct match between raw CAN data and specific OBD-II PID responses.

Each of these approaches offers a promising solution for reverse engineering vehicle CAN systems. The work presented herein builds on this prior work to further develop automated reverse engineering capabilities, enhancing both vehicle autonomy and security.

*2.3. Previous Work*

Previous work [8] proposed a methodology for reverse engineering CAN bus messages related to vehicle acceleration and deceleration controls. This method leveraged an inertial measurement unit (IMU) to capture the vehicle's physical responses to driver inputs (acceleration and deceleration) and correlated these responses with simultaneous CAN bus data recordings. By analyzing this correlation, the research identified specific CAN channels associated with the accelerator and brake pedals without needing prior knowledge of the vehicle's CAN system architecture. This approach addressed the challenge of proprietary CAN systems across different vehicle manufacturers, which lack standardized documentation and can vary in their implementation.

To validate the methodology, the research employed a systematic testing process using multiple vehicles. This involved attaching a system to each vehicle that records data from both the IMU and the CAN bus while the vehicle is operated under normal driving conditions. Post recording, the reverse engineering process included two critical phases. The first phase preprocessed the collected data to filter and separate signals indicative of acceleration and deceleration. The second phase applied a correlation analysis to identify which CAN channels' data patterns align with the IMU's recorded acceleration and deceleration events. Additionally, calibration recordings were taken with the vehicle stationary to isolate the controls' inputs, which further aided in the identification process.

The results from applying this method of reverse engineering across several vehicles demonstrated its effectiveness in identifying relevant CAN channels for accelerator and brake pedal controls without prior knowledge of the CAN system. Despite the inherent challenges posed by the proprietary nature of CAN systems and the exhaustive search required due to the lack of standardization, the research successfully identified specific CAN channels related to vehicle control inputs. These findings not only facilitate the development of aftermarket autonomous driving solutions, they also contribute to vehicle cybersecurity by enabling the monitoring of critical control messages.

## 3. Research Methodology

This section outlines the research methodology used for reverse engineering CAN messages in this study. Following the approach of previous work [8], the methodology begins by connecting a system to the vehicle to record operational data. To overcome the limitations identified in earlier research, particularly regarding deceleration correlation, this study incorporates GPS data for a more comprehensive analysis. The system captures the vehicle's movements, including GPS-based positional data, while simultaneously logging all of the CAN data generated during operation. This accumulated data undergoes processing, using the proposed techniques, to identify specific CAN channels linked to the vehicle's steering, acceleration, and braking systems.

The main components and procedures of this methodology are detailed in this section. This starts with an additional discussion of the enhancements to previous work, in section 3.1.

### 3.1. Enhancement of Previous Work

In prior work [8], it was shown that the correlation for deceleration, using a combination of IMU and CAN bus data, wasn't as strong as for acceleration. One potential reason for this is that data was collected during instances when the vehicle was already stationary and the brake pedal was still being depressed. In these cases, the vehicle could not decelerate, impairing the correlation between the brake pedal's application and deceleration. To address this issue, a new approach is proposed herein that uses GPS sensors to monitor the vehicle's speed over time. A similar method is also used for detecting the vehicle steering wheel's position and correlating and identifying the CAN channels related to the vehicle's steering wheel.

### 3.2. Method for Autonomous CAN Reverse Engineering

The proposed method begins with data being recorded while an operator drives the target vehicle. The recorded data is then post-processed and reverse engineering techniques (discussed subsequently) are used to identify which CAN channels correspond to the vehicle's controls. This paper does not present new work, with respect to acceleration, as compared to [8]; however, it does present new data and analysis regarding identifying the CAN channels for the vehicle's brake pedal and steering wheel positions. The remainder of this section discusses the benefits of the proposed approach.

**Reverse engineering vehicle steering** - The IMU sensor can also be used to identify the steering wheel position, similar to how it was used to detect accelerator and brake pedal usage in [8]. When the driver turns the steering wheel left or right, this creates linear accelerations leftwards and rightwards, much like how the accelerator and brake pedals create linear accelerations in the forward and backward directions. The accelerometer measures the magnitude of the turn, which is dependent on how far the driver turns the steering wheel and the speed of the vehicle.

**Improving correlation performance** - Eliminating the time frames when the vehicle's velocity is zero improves the correlation for deceleration, since the vehicle cannot reduce its speed further once it stops moving. To achieve this, a GPS sensor is added to the telemetry system to monitor the vehicle's velocity over time, allowing it to be recorded in parallel with the IMU and CAN data.

A similar issue also exists for steering. If the operator turns the steering wheel while the vehicle is stationary, this change cannot be measured by the IMU, impairing correlation analysis.

**Additional telemetry exploration** - Within the coordinate measurements provided by the GPS module, measurements for altitude are also included. These are not currently utilized. They could be helpful when the vehicle encounters terrain, as the vehicle may accelerate or decelerate, in these cases, without the operator using the pedals. This acceleration or deceleration, without control use, could also impair correlation analysis. Additionally, hills could also affect the raw measurements of the accelerometer as it will experience gravity in other axes, due to its new orientation. This could be, but is not currently, filtered out during the IMU preprocessing phase. Sampling the GPS data in parallel with the IMU data could help clarify if, when and where this may be needed in future research.

*3.3. System Hardware and Drivers*

This section discusses the key components and software used for the work presented herein. These are the inertial measurement unit (IMU), global positioning system (GPS), CAN analyzer tool, the Linux laptop, and the robot operating system (ROS) middleware. Figure 1 illustrates these components. This equipment is similar to that used in the previous work [8], with the addition of a GPS module.

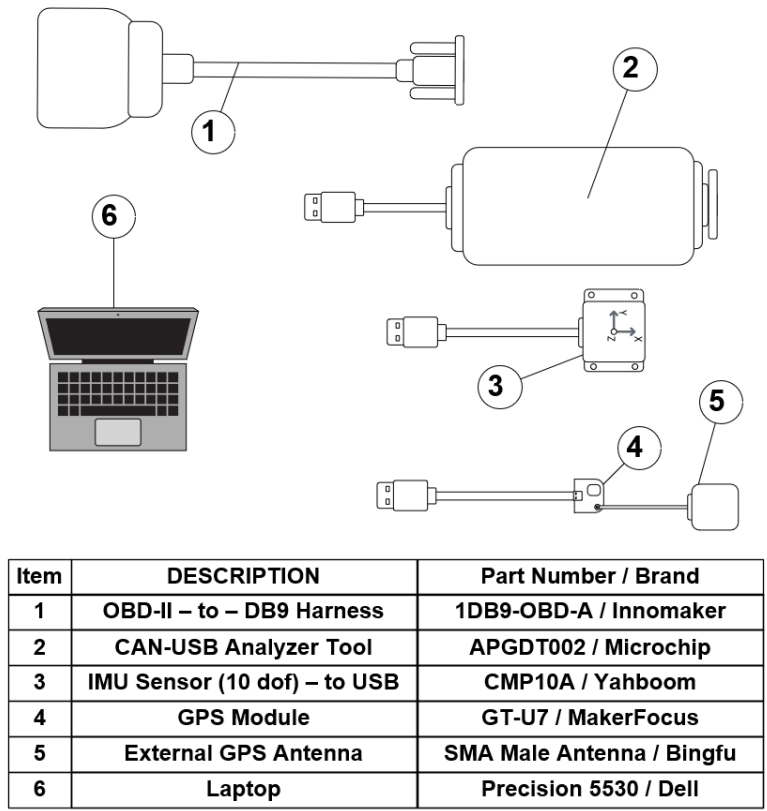

| Item | DESCRIPTION | Part Number / Brand |
|---|---|---|
| 1 | OBD-II – to – DB9 Harness | 1DB9-OBD-A / Innomaker |
| 2 | CAN-USB Analyzer Tool | APGDT002 / Microchip |
| 3 | IMU Sensor (10 dof) – to USB | CMP10A / Yahboom |
| 4 | GPS Module | GT-U7 / MakerFocus |
| 5 | External GPS Antenna | SMA Male Antenna / Bingfu |
| 6 | Laptop | Precision 5530 / Dell |

**Figure 1.** Automotive CAN reverse engineering kit (modified from **[8]**).

**Inertial measurement unit (IMU) -** The same IMU system [18] that was used in [8] was used for this research. The IMU module came with a software package that includes the necessary drivers and example software for ROS integration [19]. As before, the IMU was sampled at a rate of 100Hz. The ROS node was changed to also transmit the linear accelerations in the X-axis, and Y-axis. The IMU ROS node listens for new updates from the IMU module and, anytime a new value is presented, the IMU ROS node publishes the information to the ROS network.

**Global positioning system (GPS) -** A MakerFocus GT-U7 GPS module [20] with a micro-USB connector was used, along with an active GPS navigation antenna [21]. The GPS module outputs standard NMEA sentences, so a custom driver wasn't needed. The "*nmea_navsat_driver*" [22] ROS package includes a ROS node that receives NMEA messages and publishes data to fix and vel ROS topics. The *fix* topic provides latitude, longitude, altitude, and position covariance data. The vel topic provides velocity data for the x, y, and z axes.

**Controller area network (CAN) analyzer tool -** The Microchip ADGDT002 CAN bus analyzer tool [23] was, as in [8], used to interface with the vehicle's CAN bus. It used a USB connector, a DB9 connector and a Linux kernel driver [50]. The ROS library includes the "*socketcan_bridge*" package, which contains the "*socketcan_bridge_node*" node for translating messages between the CAN and ROS environments. CAN frames are published to the "*received_messages*" ROS topic, and ROS messages are sent to the SocketCAN [24] device through the "*sent_messages*" ROS topic. By running the *socketcan_bridge_node* [25] in the ROS environment and connecting the ADGDT002 to the vehicle CAN bus, along with instantiating the SocketCAN network stack on the device, the CAN interface is established between the ADGDT002 and the ROS network.

**Linux laptop -** A Dell Precision 5530 laptop running the Ubuntu 20.01 Linux operating system was used as both the processing system and the operator interface for the system. This laptop has an Intel Core i9 8th generation processor with 32GB of RAM. It hosts all of the software nodes for making measurements, processing data, performing reverse engineering, and presenting the information to the operator. Linux shell terminals were used to interface with the software. To maintain the visibility of all of the concurrent software during recording and analysis, the Terminator Linux application [26], [27] was used to provide multiple shell windows (threads) at once. Within Terminator, the operator opens shell windows for each of the applications and scripts within the project. Initializing the ROS environment, starting and stopping the ROS nodes, and monitoring the ROS nodes are all performed with shell scripting in Linux.

*3.4. Recording Data for Reverse Engineering*

This section discusses the process for recording the data that was used in this work. The same recordings used in [8] were also used in this study. These recordings included the collection of GPS data; however, the GPS data was not used by the methods of reverse engineering or for the analysis presented in [8]. Using these same recordings is advantageous, as it facilitates the direct characterization of the improvement that the proposed reverse engineering process provides.

**Steering wheel calibration recording –** Calibration recordings for the brake and accelerator pedals were previously collected for the subject vehicles. A steering wheel recording for calibration was collected for this study. To accomplish this, data was collected while the vehicle was running and in park. The calibration recording software was used to make a recording as the driver turned the steering wheel all the way clockwise and then all the way back counterclockwise, to the limits in each direction. This calibration recording of CAN messages during steering wheel operation is used by the vehicle controls discovery algorithm, which is described subsequently.

*3.5. Software Design Overview*

In this section, the software developed for this research is discussed. The software includes two primary algorithms.

The first, the rate of change correlation algorithm, has been designed to reverse engineer CAN channels related to acceleration, deceleration, and steering from the recorded data. Improving upon the previous work [8], this algorithm now includes a modified approach that focuses on analyzing periods when the vehicle is in motion, rather than conducting a full time series analysis as was previously done. The effectiveness of this updated method are discussed in Section 4 and analyzed in Section 5.

The second algorithm is the vehicle controls discovery algorithm. Building upon the outputs of the first, this algorithm identifies CAN channels corresponding to specific vehicle controls, incorporating the techniques and considerations discussed in subsequent subsections. Figure 2 provides an overview of the software used in this research, including the architecture for the acceleration work from [8] for contextual reference.

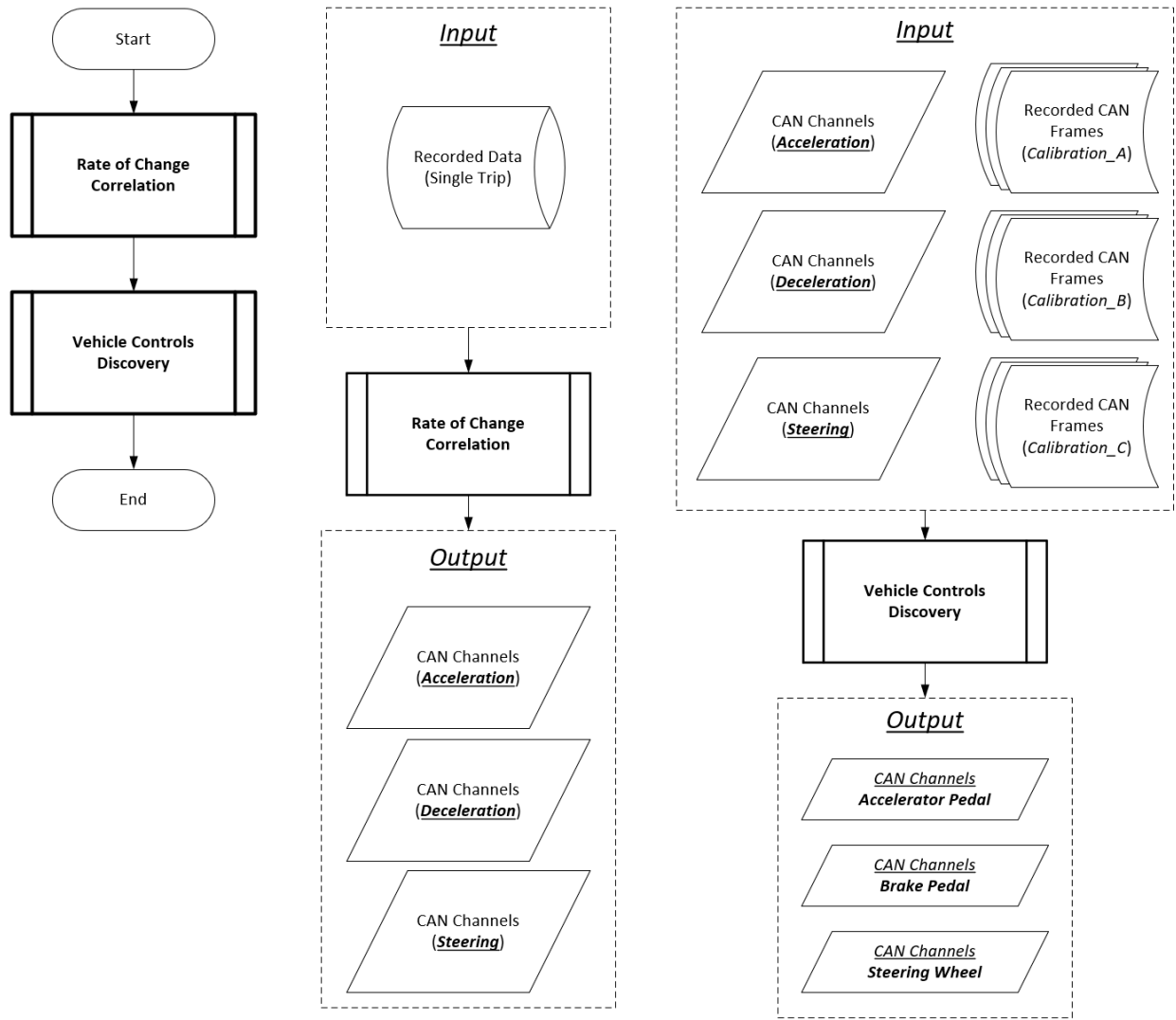

**Figure 2.** Software design overview.

**Software design: rate of change correlation -** The rate of change correlation algorithm begins by preprocessing the raw IMU data to categorize it into steering, acceleration, and deceleration measurements. Next, the GPS data is used to filter the IMU data, creating a new dataset used for correlation analysis. To filter the data with GPS, the algorithm uses the recorded GPS data to identify moments when the vehicle is stationary. These stationary points in time are then used to remove corresponding IMU samples from the dataset, leaving the data recorded during vehicle movement. This process ensures that the analysis focuses on dynamic vehicle behavior, excluding periods of inactivity.

The algorithm then uses nested FOR loops to traverse through the recorded CAN messages, breaking them down into individual CAN frames and running the Pearson correlation algorithm separately for the steering, acceleration, and deceleration data. This algorithm is depicted in Figure 3.

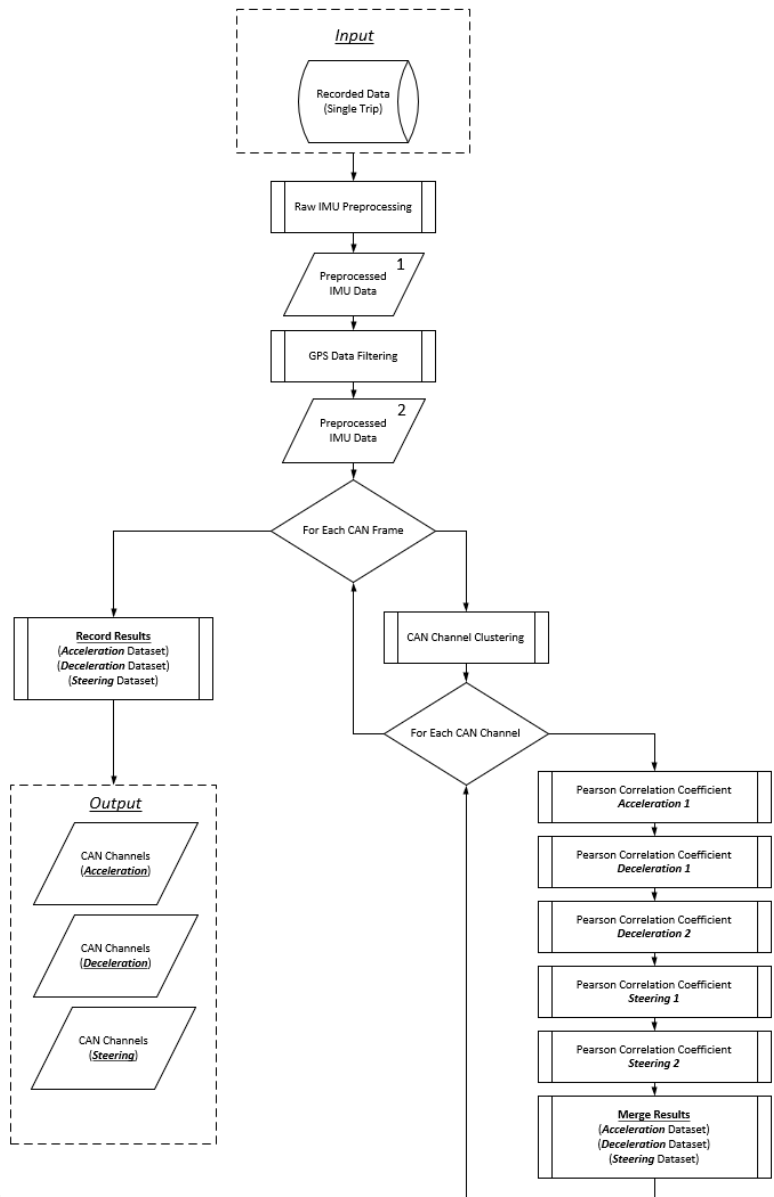

[8]

**Figure 3.** Rate of change correlation flowchart.

**Software design: vehicle controls discovery -** The vehicle controls discovery algorithm processes the highly correlated CAN channels, using data from specific calibration recordings, as discussed in section 3.4 and in previous work [8], to refine the list of CAN channels linked to each control. It examines various parameters of the data series for each CAN channel, including the directly recorded values, range, derivatives, and standard deviations, to assess their relevance and reliability in reflecting vehicle controls. Given the real-time nature of the sensor inputs used for vehicle controls, which are anticipated to be published on the CAN bus, this analysis ensures that the selected channels accurately reflect actual vehicle control inputs. The algorithm ranks channels based on the consistency of their changes over time, prioritizing those with the least variability in values. Channels that have minimal correlation and unique values are subsequently filtered out. This algorithm is illustrated in Figure 4.

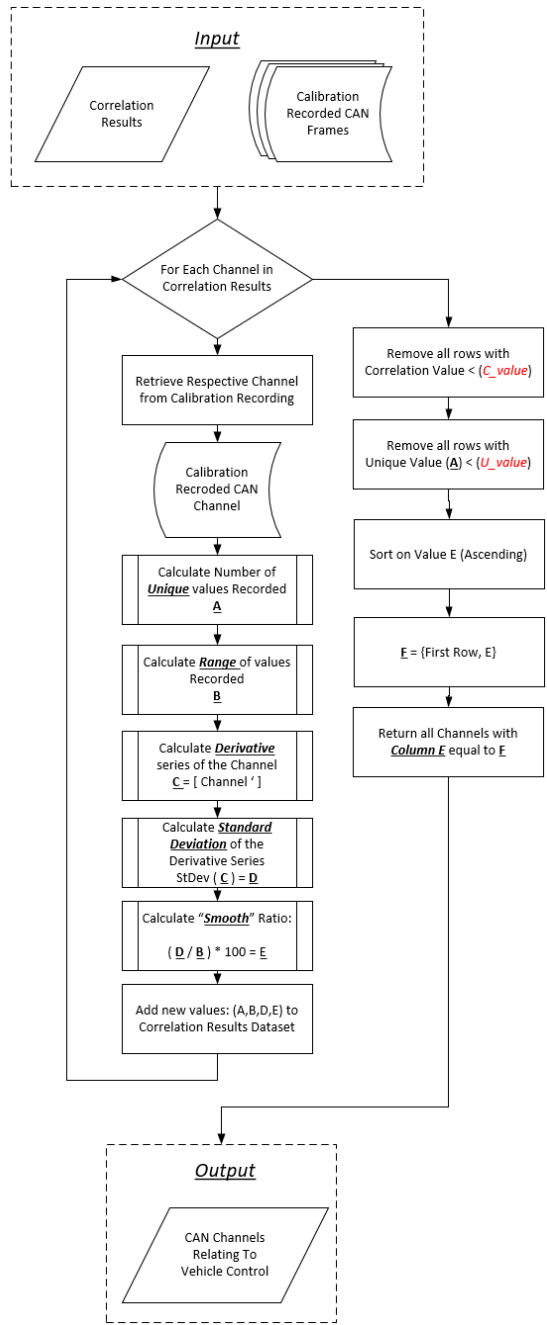

**Figure 4. Vehicle Controls Discovery Flowchart.**

## 4. Results

In this section, the data and results are presented. The subsections present the reverse engineering process and the outputs of the algorithms from recordings taken with the main test vehicle: a 2016 GMC Sierra. To assess the efficacy of this reverse engineering system, data was collected from multiple tests with this vehicle and tests with multiple additional vehicles. The results of testing with these additional vehicles are summarized at the end of this section and presented, in detail, in Appendix A.

Throughout this section, frequent reference is made to CAN channels. The terminology used for this is the same as in [8]. The generic format is: (*Frame ID)_(endianness)_(channel bit length)_*bit_*(channel number in Frame)*. Thus, "125_lsb_sixteen_bit_2" means a CAN frame ID of 125, a bit length of 16, a CAN payload channel number of 2 and an endian style of LSB. One exception to this format is that, if the bit length is eight bits (the method's default channel size), the format is simplified to: *(Frame_ID)_*byte_*(byte number)*.

### *4.1. Brake Pedal Reverse Engineering*

In this section, the reverse engineering results for the brake pedal are presented. For this data collection, the IMU sensor was orientated within the vehicle so that accelerations on the Y axis are forward and aftward accelerations of the vehicle. Vehicle deceleration was represented by positive y-axis values and acceleration produced negative y-axis values. The Y-axis of the IMU sensor was filtered to generate two distinct signals: one for acceleration and one for deceleration. Figure 5 presents the results for a recorded trip using the main test vehicle, a 2016 GMC Sierra. The left graph is the deceleration recording without considering the GPS data and the right graph is the deceleration recording with the GPS data considered. Whenever the GPS indicated that the vehicle's velocity was zero, the recorded data was removed from analysis. For example, between roughly 110 seconds and 140 seconds, in Figure 5, the GPS indicated that the vehicle was at rest and, thus, the data within that time frame was removed from analysis.

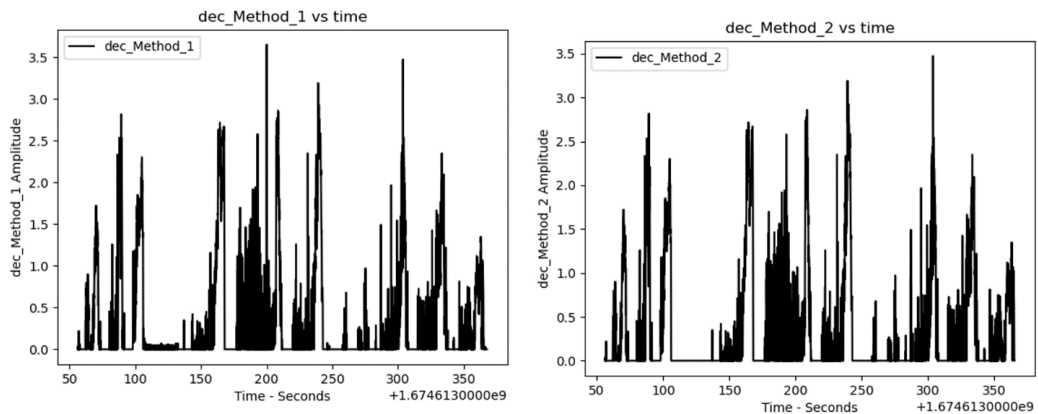

**Figure 5.** Vehicle acceleration and deceleration values measured by the IMU during a recorded trip.

The rate of change correlation algorithm used the deceleration signal as a baseline for correlation with all CAN channels recorded during the drive. For every identified CAN channel, a correlation (presented as a decimal value to represent a fraction of 1.0) was calculated to determine how closely the specific recorded CAN channel related to vehicle deceleration. The top 25 results from this correlation are shown in Table 1.

This information was supplied to the vehicle controls discovery algorithm, which narrowed down the CAN channels associated with deceleration to only those that are likely to be linked to the brake pedal. The vehicle controls discovery algorithm processes data from the deceleration correlation results, focusing on CAN channels associated with the brake pedal. To accomplish this, the algorithm utilizes the results from the rate of change correlation algorithm; however, it also conducts analysis using the brake pedal calibration recording instead of the vehicle drive recording.

Table 1. Deceleration results with GPS.

| ID | Channel | Correlation | ID | Channel | Correlation |
|---|---|---|---|---|---|
| 532 | msb_fourteen_bit_1 | 0.90920095 | 761 | msb_fourteen_bit_5 | 0.88938133 |
| 761 | msb_thirteen_bit_5 | 0.90890751 | 209 | msb_fourteen_bit_0 | 0.8779284 |
| 209 | msb_eleven_bit_0 | 0.90585026 | 209 | msb_thirteen_bit_0 | 0.8779284 |
| 508 | msb_eleven_bit_4 | 0.90312916 | 209 | msb_twelve_bit_0 | 0.8779284 |
| 510 | msb_twelve_bit_0 | 0.90098864 | 508 | byte_4 | 0.87659406 |
| 510 | msb_ten_bit_1 | 0.89994872 | 508 | msb_ten_bit_3 | 0.87659406 |
| 508 | msb_fifteen_bit_4 | 0.89463714 | 508 | msb_nine_bit_3 | 0.87659406 |
| 508 | msb_twelve_bit_4 | 0.89463714 | 532 | msb_sixteen_bit_1 | 0.87502252 |
| 508 | msb_thirteen_bit_4 | 0.89463714 | 532 | byte_1 | 0.87499889 |
| 508 | msb_sixteen_bit_4 | 0.89463714 | 532 | msb_nine_bit_0 | 0.87499889 |
| 508 | msb_fourteen_bit_4 | 0.89463714 | 532 | lsb_nine_bit_1 | 0.87499889 |
| 761 | lsb_eleven_bit_5 | 0.89292913 | 761 | msb_fifteen_bit_5 | 0.8742779 |
| 532 | msb_fifteen_bit_1 | 0.8898991 | | | |

The algorithm augments the dataset by adding calculated attributes to each CAN channel, which are indicated in Table 2's additional columns: 'Range', 'Unique', 'StDev(*)', and 'Smooth'. The algorithm evaluates each channel's activity during the brake pedal calibration, determines the range of values (displayed in the 'Range' column) and counts the number of unique values (which is listed in the 'Unique' column). It also calculates a derivative series from the recorded time series for each CAN channel and the standard deviation of this series. These results are placed in the 'StDev(*)' column. The 'Smooth' column value is determined by dividing the 'StDev(*)' value by the 'Range' value and then multiplying it by 100 to convert it into a percentage. Finally, the algorithm orders the data series based on the 'Smooth' column in ascending order. CAN channels with values equal to the smallest increment in this column are selected for use. In Table 2, channels with a 'Smooth' value below one indicate that the CAN channel data changes within 1% of its potential range, highlighting the sensitivity of these channels during brake pedal engagement.

Table 2. 2016 GMC Sierra 1500 potential brake pedal channels with GPS.

| ID | Channel | Correlation | Range | Unique | StDev(*) | Smooth |
|---|---|---|---|---|---|---|
| 209 | msb_fourteen_bit_0 | 0.87792839 | 5447 | 481 | 26 | 1 |
| 209 | msb_thirteen_bit_0 | 0.87792839 | 5447 | 481 | 26 | 1 |
| 190 | lsb_sixteen_bit_0 | 0.86175664 | 42508 | 368 | 253 | 1 |
| 190 | msb_sixteen_bit_1 | 0.8552687 | 42496 | 157 | 315 | 1 |
| 241 | byte_1 | 0.8526338 | 129 | 128 | 1 | 1 |
| 241 | lsb_twelve_bit_1 | 0.8526338 | 129 | 128 | 1 | 1 |
| 241 | lsb_thirteen_bit_1 | 0.8526338 | 129 | 128 | 1 | 1 |
| 241 | lsb_eleven_bit_1 | 0.8526338 | 129 | 128 | 1 | 1 |
| 241 | lsb_ten_bit_1 | 0.8526338 | 129 | 128 | 1 | 1 |
| 241 | lsb_nine_bit_1 | 0.8526338 | 129 | 128 | 1 | 1 |
| 241 | msb_nine_bit_0 | 0.8526338 | 129 | 128 | 1 | 1 |
| 241 | msb_sixteen_bit_1 | 0.85261112 | 33028 | 356 | 186 | 1 |
| 241 | lsb_sixteen_bit_0 | 0.6509576 | 33150 | 377 | 246 | 1 |

The most highly correlated CAN channels with the brake pedal were determined and are presented in Table 2. Figures 6 and 7 Figure 7illustrate two of these channels that clearly correspond to the brake pedal's position.

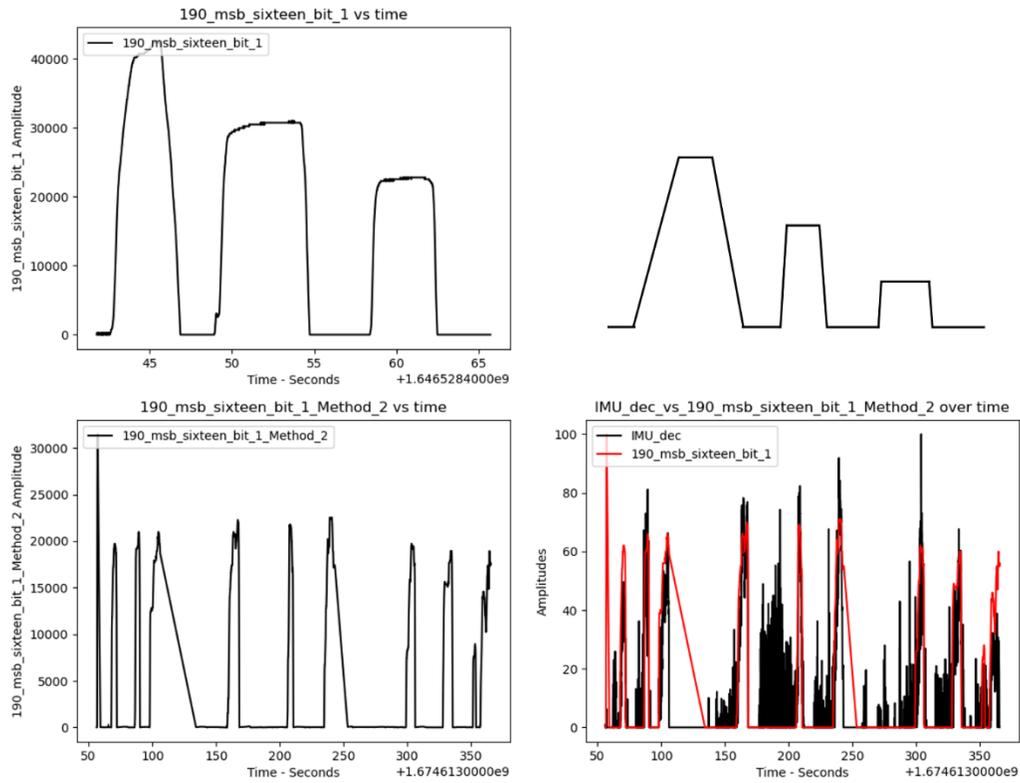

**Figure 6.** 2016 GMC Sierra 1500 potential brake pedal channel: 190 Sixteen-bit (msb) channel 1 with GPS.

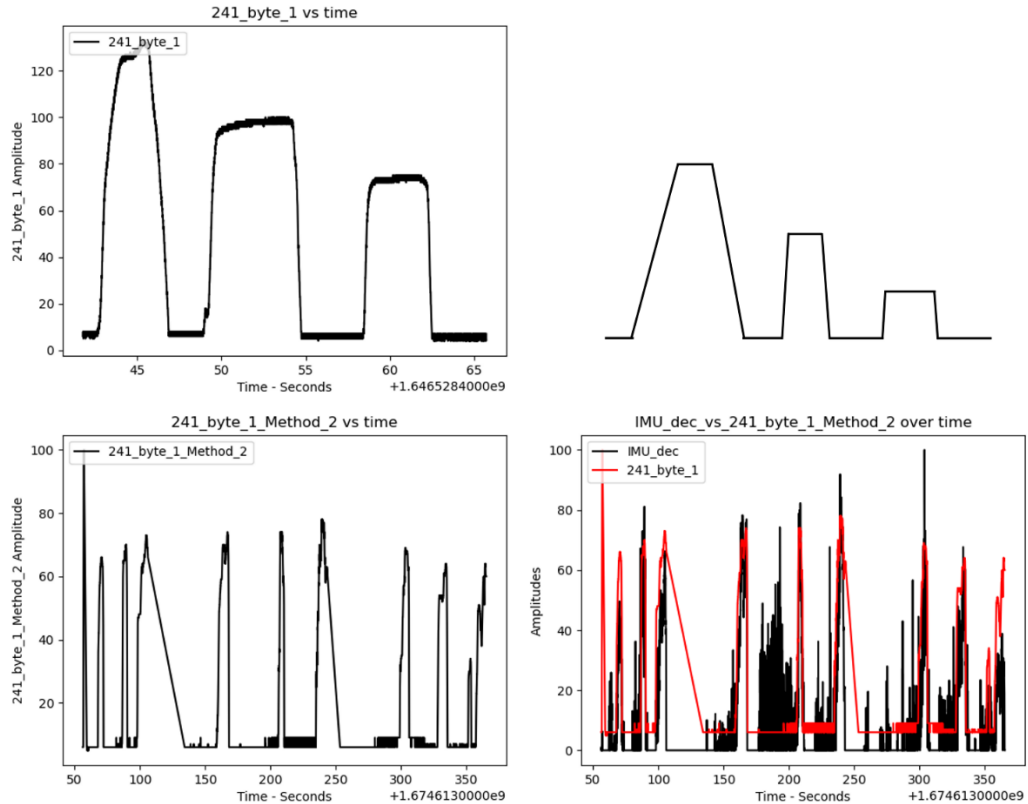

**Figure 7.** 2016 GMC Sierra 1500 brake pedal position channel 241_byte_1 with GPS. Within each figure, the top images display the CAN channel data during the calibration recording of the brake pedal and the lower images show the CAN channel data collected during the vehicle trip recording. The lower images use the new method, presented in this work, which incorporated GPS data along with the IMU data to eliminate the times where the vehicle was at rest. Figures 6 and 7 illustrate that data was removed at the periods in time where the vehicle's velocity was zero. Examples of this are between approximately 110 and 140 seconds and at approximately 250 seconds.

*4.2. Steering Wheel Reverse Engineering*

In this section, the reverse engineering results for the brake pedal are presented. For this data collection, the IMU sensor was orientated within the vehicle so that accelerations on the X axis correspond to the vehicle's leftwards and rightwards movements. The vehicle's leftwards acceleration was represented as positive X-axis values and rightwards acceleration was represented as negative X-axis values.

Figure 8 shows data for a recorded trip using the main test vehicle: a 2016 GMC Sierra. The left image is the steering recording without GPS data being considered and the right image is the steering recording with GPS data utilized. Whenever the GPS identified that the vehicle's velocity was zero, the recorded data was removed from analysis. Examples of this are between 110 and 140 seconds, and again at approximately 250 seconds.

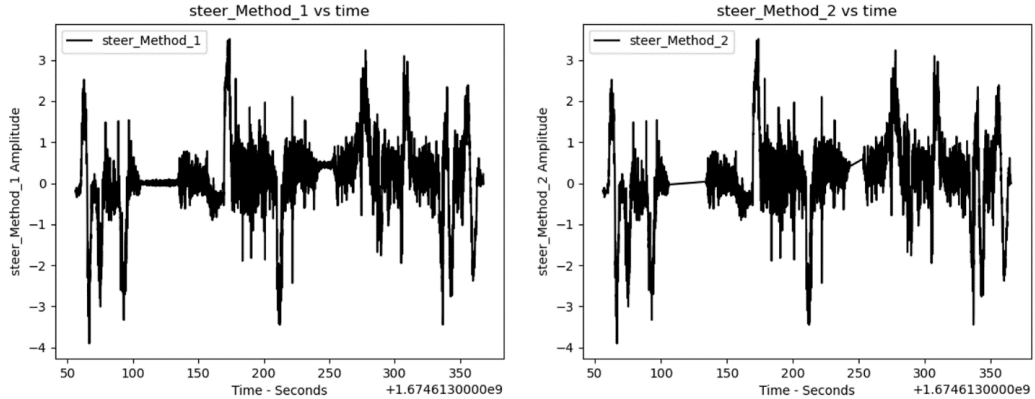
**Figure 8.** Vehicle steering results measured by the IMU during a recorded trip.

The rate of change correlation algorithm used the steering signal values recorded on the IMU X axis as a baseline for correlation with all of the CAN channels recorded during the drive. For every identified CAN channel, a correlation value was calculated to indicate how closely the CAN channel related to vehicle steering. The top 25 results from each correlation method (with and without GPS) are shown in Tables 3 and 4.

**Table 3.** Steering results without GPS.

| ID | Channel | Correlation | ID | Channel | Correlation |
|----|---------|-------------|----|---------|-------------|
| 564 | lsb_nine_bit_2 | 0.715809247 | 844 | msb_twelve_bit_6 | 0.582025089 |
| 564 | byte_2 | 0.715809247 | 844 | msb_thirteen_bit_6 | 0.582025089 |
| 564 | msb_fifteen_bit_2 | 0.715731029 | 844 | msb_sixteen_bit_6 | 0.582025089 |
| 564 | msb_sixteen_bit_2 | 0.715731029 | 508 | msb_fifteen_bit_2 | 0.576772718 |
| 844 | byte_6 | 0.598746279 | 489 | msb_eleven_bit_0 | 0.570214331 |
| 508 | byte_2 | 0.591221043 | 844 | msb_eleven_bit_6 | 0.56432174 |
| 508 | msb_sixteen_bit_2 | 0.590850914 | 508 | msb_fourteen_bit_2 | 0.546598642 |
| 489 | msb_fourteen_bit_0 | 0.587803728 | 844 | byte_2 | 0.541213005 |
| 489 | msb_twelve_bit_0 | 0.587803728 | 489 | msb_fifteen_bit_3 | 0.5405035 |
| 489 | msb_thirteen_bit_0 | 0.587803728 | 489 | msb_eleven_bit_3 | 0.5405035 |
| 844 | msb_fourteen_bit_6 | 0.582025089 | 489 | msb_twelve_bit_3 | 0.5405035 |
| 844 | msb_fifteen_bit_6 | 0.582025089 | 489 | msb_sixteen_bit_3 | 0.5405035 |

**Table 4.** Steering results with GPS.

| ID | Channel | Correlation | ID | Channel | Correlation |
|----|---------|-------------|----|---------|-------------|
| 564 | byte_2 | 0.71924371 | 844 | msb_twelve_bit_6 | 0.60271833 |
| 564 | lsb_nine_bit_2 | 0.71924371 | 844 | msb_sixteen_bit_6 | 0.60271833 |
| 564 | msb_fifteen_bit_2 | 0.7191718 | 844 | msb_fourteen_bit_6 | 0.60271833 |
| 564 | msb_sixteen_bit_2 | 0.7191718 | 508 | msb_fifteen_bit_2 | 0.5975437 |
| 844 | byte_6 | 0.61919651 | 489 | msb_eleven_bit_0 | 0.59131281 |
| 508 | byte_2 | 0.61170921 | 844 | msb_eleven_bit_6 | 0.58520244 |
| 508 | msb_sixteen_bit_2 | 0.61144779 | 508 | msb_fourteen_bit_2 | 0.56759151 |
| 489 | msb_twelve_bit_0 | 0.60869945 | 844 | byte_2 | 0.55331612 |
| 489 | msb_fourteen_bit_0 | 0.60869945 | 489 | msb_ten_bit_0 | 0.55322086 |
| 489 | msb_thirteen_bit_0 | 0.60869945 | 489 | msb_fifteen_bit_3 | 0.55268058 |
| 844 | msb_thirteen_bit_6 | 0.60271833 | 489 | msb_twelve_bit_3 | 0.55268058 |
| 844 | msb_fifteen_bit_6 | 0.60271833 | 489 | msb_fourteen_bit_3 | 0.55268058 |

The vehicle controls discovery algorithm uses this information and narrows down the CAN channels associated with movement to only those that are likely to be associated with the steering wheel. The vehicle controls discovery algorithm processes data from the steering correlation results, focusing on CAN channels associated with the steering mechanism. This is achieved by using the results from the rate of change correlation algorithm, with further analysis conducted using the steering wheel calibration recording rather than the vehicle drive recording.

The proposed method augments the dataset by appending calculated attributes to each CAN channel, as shown in Tables 5 and 6. These calculated values are 'Range', 'Unique', 'StDev(*)', and 'Smooth'. The algorithm evaluates each channel's behavior during the steering wheel calibration and determines the range of values (shown in the 'Range' column) and the number of unique values (listed in the 'Unique' column). It also calculates a derivative series from the time series recorded for each CAN channel. Then, the standard deviation of this series is computed and is listed in the 'StDev(*)' column. The 'Smooth' column value is then derived by dividing the 'StDev(*)' value by the 'Range' value and multiplying by 100 to express it as a percentage. Finally, the algorithm arranges the series based on the 'Smooth' column in ascending order. The CAN channels whose values match the smallest increment in this column are selected for use. This data is shown in Tables 5 and 6. Channels with a 'Smooth' value below one indicate that the CAN channel data changes within 1% of its potential range, emphasizing the sensitivity of these channels during steering wheel engagement.

**Table 5.** 2016 GMC Sierra 1500 potential steering wheel position channels without GPS.

| ID | Channel | Correlation | Range | Unique | StDev(*) | Smooth |
|---|---|---|---|---|---|---|
| 564 | lsb_nine_bit_2 | 0.71580925 | 113 | 102 | 1 | 1 |
| 564 | byte_2 | 0.71580925 | 113 | 102 | 1 | 1 |
| 564 | msb_fifteen_bit_2 | 0.71573103 | 28928 | 102 | 142 | 1 |
| 564 | msb_sixteen_bit_2 | 0.71573103 | 28928 | 102 | 142 | 1 |

The most highly correlated CAN channels with the brake pedal were identified and are presented in Table 5 (for the first method, without GPS) and Table 6 (for the second method, with GPS). Figures 9 and 10 illustrate one of these channels that clearly corresponds to the brake pedal.

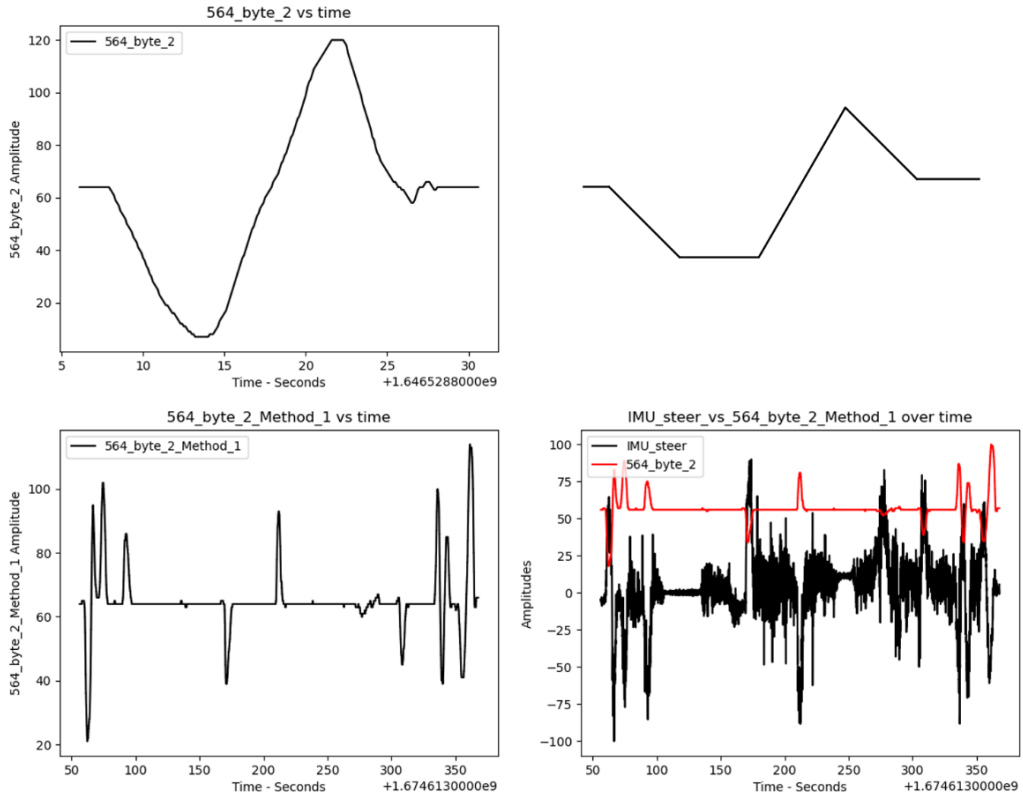

**Figure 9.** 2016 GMC Sierra 1500 potential steering wheel channel: 564 byte 2 without GPS.

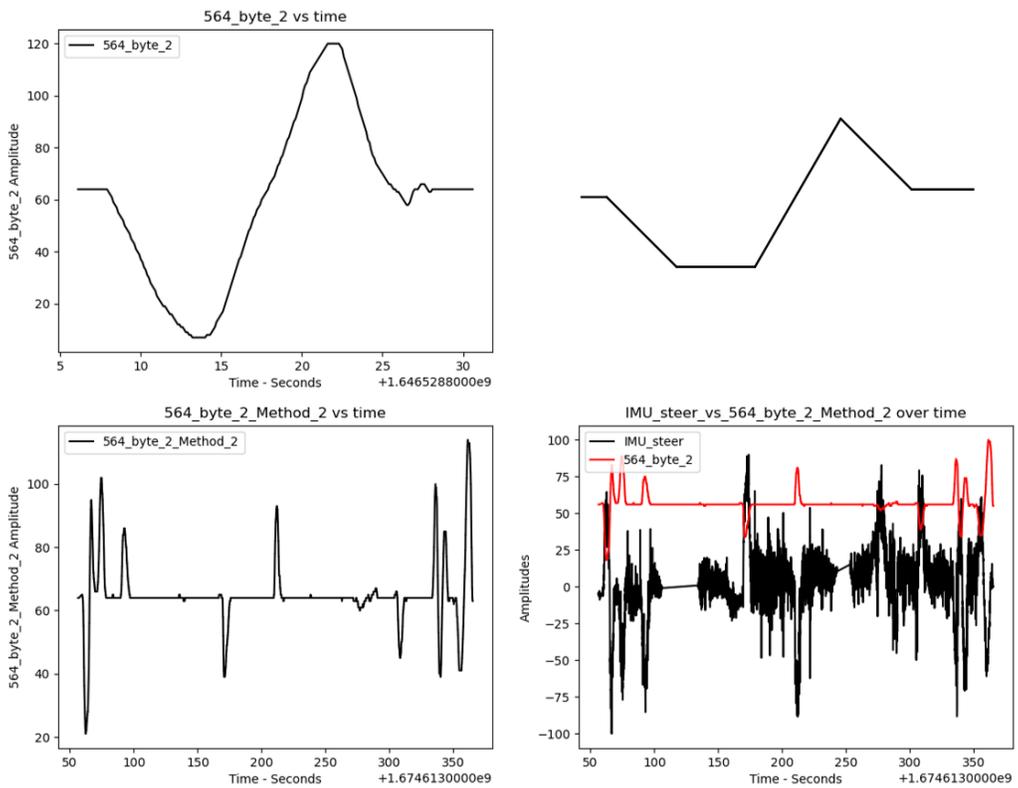

**Figure 10.** 2016 GMC Sierra 1500 potential steering wheel channel: 564 byte 2 with GPS.

Table 6. 2016 GMC Sierra 1500 potential steering wheel position channels with GPS.

| ID | Channel | Correlation | Range | Unique | StDev(*) | Smooth |
|---|---|---|---|---|---|---|
| 564 | byte_2 | 0.71924371 | 113 | 102 | 1 | 1 |
| 564 | lsb_nine_bit_2 | 0.71924371 | 113 | 102 | 1 | 1 |
| 564 | msb_fifteen_bit_2 | 0.7191718 | 28928 | 102 | 142 | 1 |
| 564 | msb_sixteen_bit_2 | 0.7191718 | 28928 | 102 | 142 | 1 |

In both figures, the upper images display the CAN channel data during the calibration recording of the brake pedal and the lower images show the CAN channel data during the vehicle trip recording. The lower images in Figure 9 show the use of the first method (without GPS) and the lower images in Figure 10 show the use of the second method (with GPS).

**5. Discussion**

This section presents a discussion of the research results. It begins with the interpretation of the findings, providing an analysis of the outcomes. Subsequently, the results are compared to previous research, highlighting the advancements made and the context within which this study contributes to automotive CAN reverse engineering.

This section presents conclusions from the research findings and identifies directions for future studies in this area. The implications of the current study's contributions and how they might influence ongoing and future research endeavors are discussed.

*5.1. Interpretation and Analysis of Results*

This section provides analysis of the results obtained from the proposed reverse engineering process. It focuses on the correlation of CAN channels to vehicle actions and the identification of CAN channels related to specific vehicle controls. The enhancements made in this study, over previous work, are also examined.

5.1.1. Correlating CAN Channels to Vehicle Actions

Similar to the findings in the previous work [8], this study has demonstrated strong correlations between IMU data and CAN channels. Notably, the integration of GPS data has refined these correlations, particularly for deceleration actions, enhancing the precision of the reverse engineering process. This study also used IMU's X-axis data to reverse engineer steering actions. The methodology, like that used for acceleration and deceleration with the IMU's Y-axis, proved effective. Most vehicles tested, except for the Impala, successfully identified CAN channels strongly correlated with steering wheel position. In the primary vehicle, the 2016 GMC Sierra, the inclusion of GPS data led to a modest improvement of about 1% in steering results. This aligned with expectations that steering actions predominantly occurred during vehicle motion in the specified recording.

5.1.2. Identifying CAN Channels Relating to Vehicle Controls

The algorithms in this research have effectively identified CAN channels related to the accelerator [8] and brake pedals and, now, the steering wheel. By analyzing the collected CAN channels associated with vehicle acceleration, deceleration and steering, relevant CAN channels for each vehicle tested were successfully identified. The algorithm performed consistently across all vehicles, though the specific CAN channels that were identified varied among different manufacturers. Among GM and Chevy vehicles, the results showed a high degree of similarity.

A key observation in the steering analysis was the identification of only one arbitration ID. This suggests that, among the four channels identified, only one likely represents the true steering wheel channel. This issue, which is indicative of potential overlaps and ambiguities in CAN channel tokenization, was also experienced in deceleration and acceleration analyses. For example, in the 2016 GMC Sierra, four results for the steering wheel position were found within CAN arbitration ID 564. This indicates a need for further refinement in signal boundary identification, potentially by integrating methods like the READ algorithm [14], to determine the correct signals more accurately.

5.1.3. Comparison to Previous Work

The enhancements in this study, particularly the use of GPS data, have led to significant improvements in deceleration correlation values, as evidenced in the primary vehicle, the 2016 GMC Sierra. The correlation for 'channel 241_byte_1', a strong brake pedal channel candidate, increased from 55.8% to 85.3%, as compared to previous results [8]. However, it's important to note that these improvements were not uniform across all tested vehicles. Recordings lacking extended stops at intersections showed less or no improvement, highlighting the variability in correlation enhancements across different driving scenarios, when not using the GPS data.

*5.2. Limitations and Recommendations for Future Research*

This study introduced a new method, using GPS data, which improved the correlation performance for deceleration CAN channel identification. It also introduced a method for steering wheel position reverse engineering. Improving the deceleration identification addressed a key issue that was identified in [8]; however, other limitations discussed in previous work were not addressed in this study. Several of these areas remain key topics for future work.

These limitations include the large number of possible CAN channel combinations for each vehicle tested and the lack of being able to perform this analysis in real-time. Previous studies (e.g., [12], [14], [30]) have proposed techniques for signal identification which would eliminate the need for exhaustive searches for CAN channel lengths. Future work could incorporate these techniques or explore other signal boundary identification methods. Improvements to the IMU and GPS system, for measuring vehicle actions and additional methods of filtering and tracking, could be also explored, potentially using machine learning techniques. Despite the current limitations, this system can successfully reverse engineer the accelerator pedal, brake pedal, and steering wheel CAN channels without prior knowledge of a vehicle's CAN bus data format.

**6. Conclusions and future work**

This paper has described a method for reverse engineering a vehicle's steering wheel position CAN channels without any prior knowledge of the vehicle's CAN bus data format. It utilizes IMU and GPS data to detect vehicle actions and match them with specific CAN channels. The results presented herein demonstrated that incorporating GPS data greatly enhances the correlation values for deceleration and, specifically, for identifying the brake pedal CAN channels of a 2016 GMC Sierra.

Additionally, using IMU measurements to determine the vehicle's lateral movement was shown to be a successful method for creating an inertial dataset that can be matched with steering CAN channels. This allowed for the identification of CAN channels related to vehicles' steering wheel position using the algorithms presented herein.

Despite the limitations of this study, including the large number of possible CAN channel combinations for each vehicle and the requirement for conducting analysis using post-processing, this system successfully reverse engineered the accelerator pedal, brake pedal, and steering wheel CAN channels without any prior knowledge of the vehicles tested. This work can be used to support the development of an aftermarket autonomous driving capability for vehicles, and to improve vehicle security and threat analysis.

The ability to identify specific CAN messages related to vehicle controls without requiring prior information about the vehicle can be a valuable tool for reverse engineering and for researchers working to improve vehicle security alike. Future research can focus on addressing the limitations of this study, such as exploring alternative signal boundary identification methods and improving the IMU and GPS system for measuring vehicle actions. Overall, the presented approach has demonstrated a promising avenue for further research in the field of automotive CAN reverse engineering.

20 of 36

# Appendix

In this appendix, the brake pedal and steering wheel position data and results are presented for several additional vehicles: a 2021 GMC Sierra 2500, a 2022 Chevrolet Traverse, a 2009 Chevrolet Impala, a 2006 Volvo XC90, and a 2016 Ford Fusion. Each vehicle is presented in a dedicated subsection within this appendix.

*A.1. 2021 GMC Sierra 2500 Results*

This section presents the data for the 2021 GMC Sierra 2500. Table A1 lists the best correlating CAN channels for the brake pedal. Figure A1 illustrates the results for CAN channel 241_byte_1 and Figure A2 illustrates the results for CAN channel 241_byte_4. For each figure, the top images show the CAN channel data during the accelerator pedal calibration recording. The bottom images show the CAN channel data during the vehicle trip recording.

**Table A1.** 2021 GMC Sierra 2500 results brake pedal position with GPS.

| ID | Channel | Correlation | Range | Unique | StDev(*) | Smooth |
|---|---|---|---|---|---|---|
| 190 | msb_sixteen_bit_1 | 0.76525773 | 27648 | 108 | 264 | 1 |
| 190 | msb_fifteen_bit_1 | 0.76525773 | 27648 | 108 | 264 | 1 |
| 190 | lsb_sixteen_bit_0 | 0.765164599 | 27660 | 321 | 243 | 1 |
| 241 | lsb_sixteen_bit_0 | 0.764614809 | 30078 | 385 | 210 | 1 |
| 241 | msb_nine_bit_3 | 0.764258181 | 234 | 118 | 2 | 1 |
| 241 | msb_fourteen_bit_3 | 0.764258181 | 234 | 118 | 2 | 1 |
| 241 | msb_ten_bit_3 | 0.764258181 | 234 | 118 | 2 | 1 |
| 241 | msb_twelve_bit_3 | 0.764258181 | 234 | 118 | 2 | 1 |
| 241 | msb_thirteen_bit_3 | 0.764258181 | 234 | 118 | 2 | 1 |
| 241 | msb_eleven_bit_3 | 0.764258181 | 234 | 118 | 2 | 1 |
| 241 | msb_sixteen_bit_3 | 0.764258181 | 234 | 118 | 2 | 1 |
| 241 | lsb_twelve_bit_1 | 0.764258181 | 117 | 118 | 1 | 1 |
| 241 | byte_4 | 0.764258181 | 234 | 118 | 2 | 1 |
| 241 | lsb_fourteen_bit_1 | 0.764258181 | 117 | 118 | 1 | 1 |
| 241 | msb_nine_bit_0 | 0.764258181 | 117 | 118 | 1 | 1 |
| 241 | lsb_fifteen_bit_1 | 0.764258181 | 117 | 118 | 1 | 1 |
| 241 | lsb_eleven_bit_1 | 0.764258181 | 117 | 118 | 1 | 1 |
| 241 | lsb_ten_bit_1 | 0.764258181 | 117 | 118 | 1 | 1 |
| 241 | msb_fifteen_bit_3 | 0.764258181 | 234 | 118 | 2 | 1 |
| 241 | lsb_thirteen_bit_1 | 0.764258181 | 117 | 118 | 1 | 1 |
| 241 | byte_1 | 0.764258181 | 117 | 118 | 1 | 1 |
| 241 | lsb_sixteen_bit_3 | 0.764258181 | 59904 | 118 | 422 | 1 |
| 241 | lsb_nine_bit_1 | 0.764258181 | 117 | 118 | 1 | 1 |
| 241 | msb_sixteen_bit_4 | 0.764257961 | 59907 | 416 | 434 | 1 |
| 241 | msb_fifteen_bit_1 | 0.764257899 | 29952 | 192 | 216 | 1 |

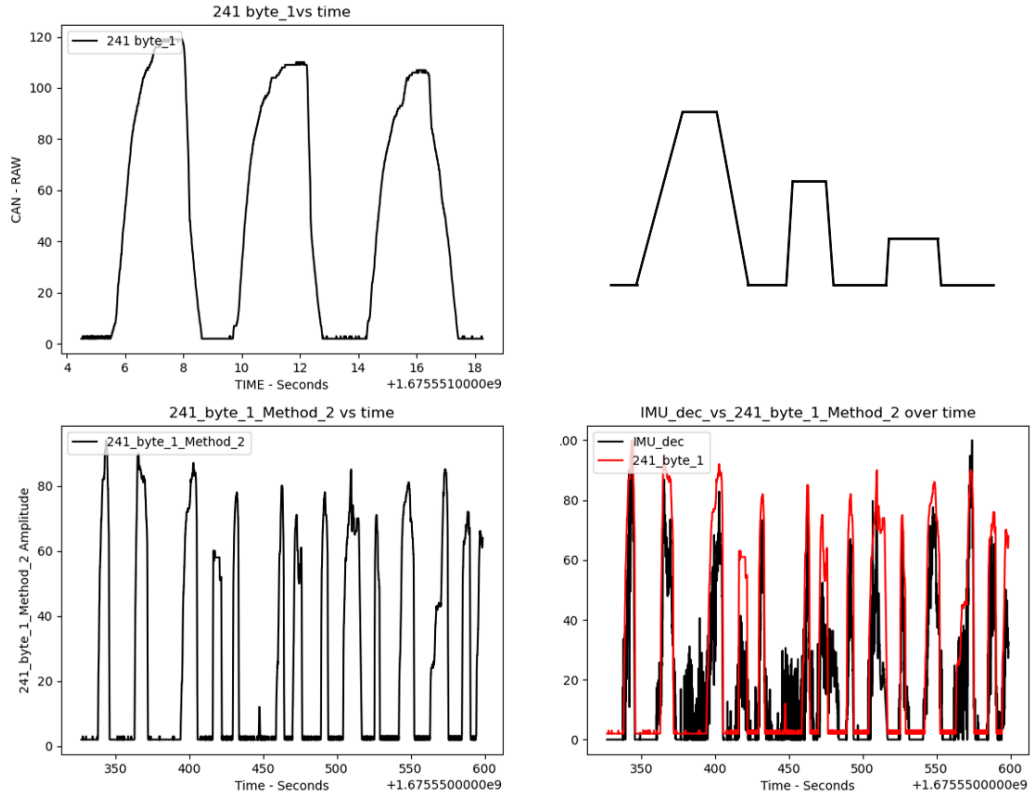

**Figure A1.** 2021 GMC Sierra brake pedal position channel 241_byte_1 with GPS.

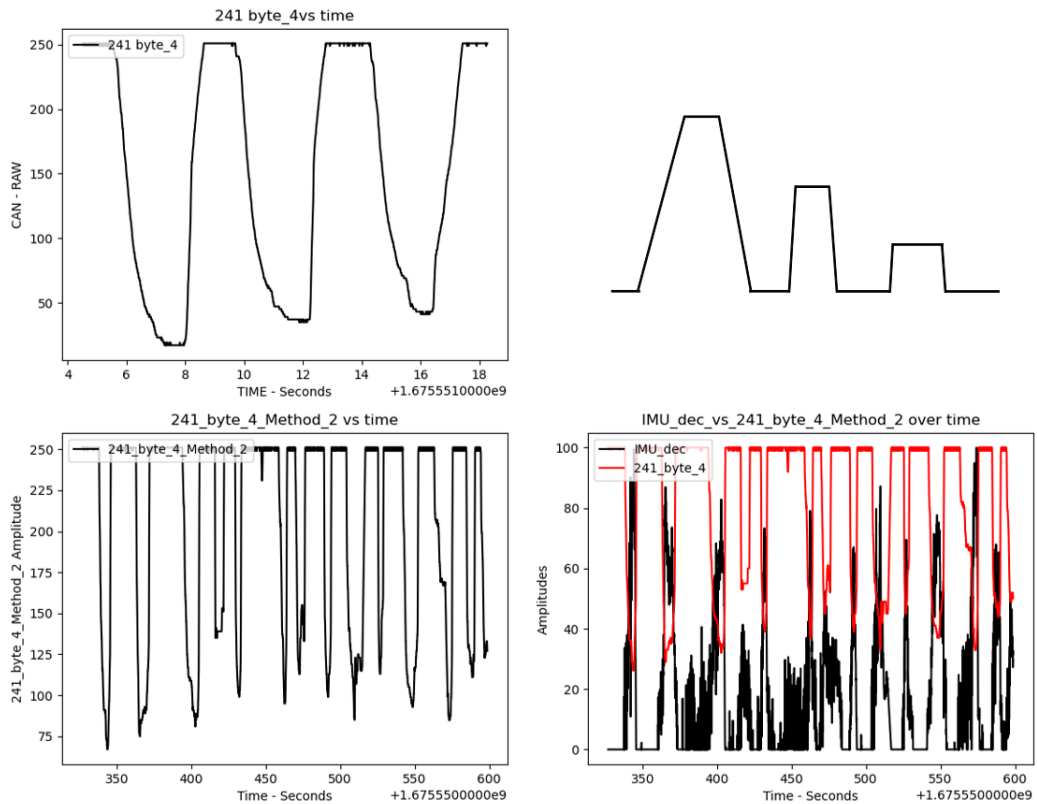

**Figure A2.** 2021 GMC Sierra brake pedal position channel 241_byte_4 with GPS.

Table A2 lists the best correlating CAN channels for the steering wheel position found with the first method (without GPS data), for the 2021 GMC Sierra 2500. Figure A3 presents the results for CAN channel 564_byte_2. The top images show the CAN channel data during the accelerator pedal calibration recording. The bottom images show the CAN channel data during the vehicle trip recording.

**Table A2.** 2021 GMC Sierra 2500 results steering wheel position without GPS.

| ID | Channel | Correlation | Range | Unique | StDev(*) | Smooth |
|---|---|---|---|---|---|---|
| 564 | msb_sixteen_bit_2 | 0.820791315 | 28672 | 85 | 174 | 1 |
| 564 | msb_fifteen_bit_2 | 0.820791315 | 28672 | 85 | 174 | 1 |
| 564 | lsb_nine_bit_2 | 0.820790742 | 112 | 85 | 1 | 1 |
| 564 | byte_2 | 0.820790742 | 112 | 85 | 1 | 1 |
| 564 | lsb_sixteen_bit_1 | 0.820789636 | 28672 | 85 | 174 | 1 |

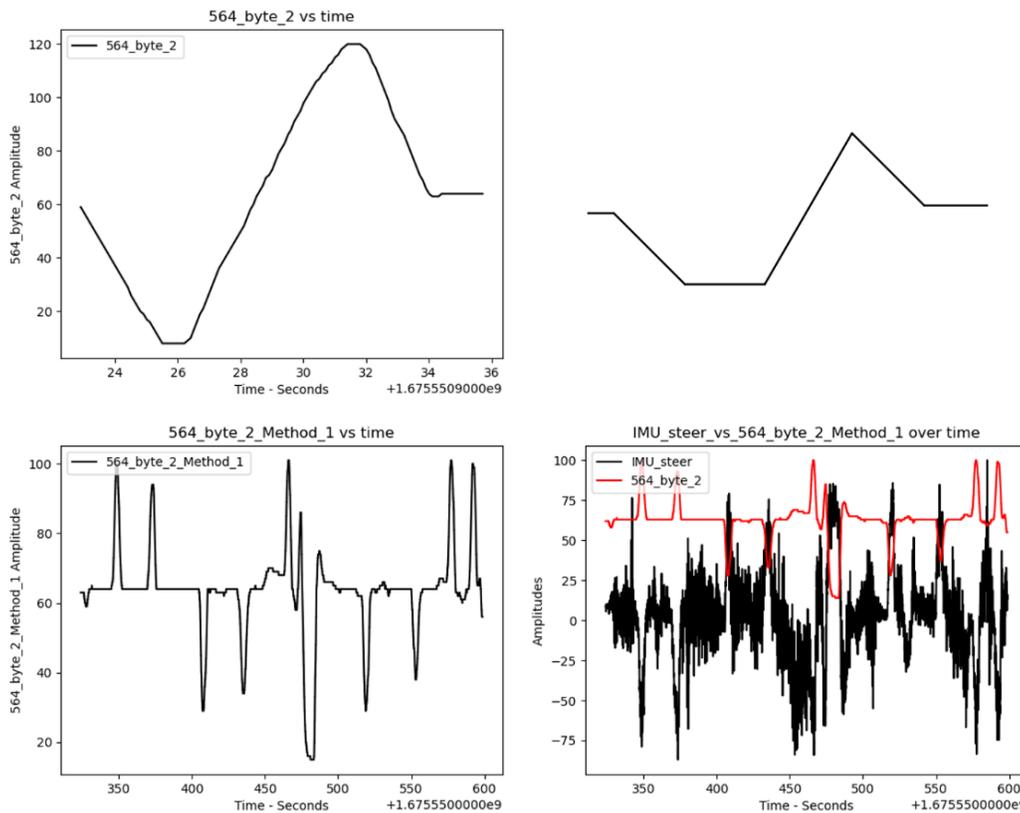

**Figure A3.** 2021 GMC Sierra 2500 steering wheel position verification: 564 byte 2 without GPS.

Table A3 lists the best correlating CAN channels for the steering wheel position found using the second method (using GPS data) for the 2021 GMC Sierra 2500. Figure A4 presents the results for CAN channel 564_byte_2. The top images show the CAN channel data during the accelerator pedal calibration recording. The bottom images show the CAN channel data during the vehicle trip recording.

**Table A3.** 2021 GMC Sierra 2500 results steering wheel position with GPS.

| ID | Channel | Correlation | Range | Unique | StDev(*) | Smooth |
|---|---|---|---|---|---|---|
| 564 | msb_sixteen_bit_2 | 0.83179409 | 28672 | 85 | 174 | 1 |
| 564 | msb_fifteen_bit_2 | 0.83179409 | 28672 | 85 | 174 | 1 |

| 564 | lsb_sixteen_bit_1 | 0.831785476 | 28672 | 85 | 174 | 1 |
| 564 | byte_2 | 0.831784822 | 112 | 85 | 1 | 1 |
| 564 | lsb_nine_bit_2 | 0.831784822 | 112 | 85 | 1 | 1 |

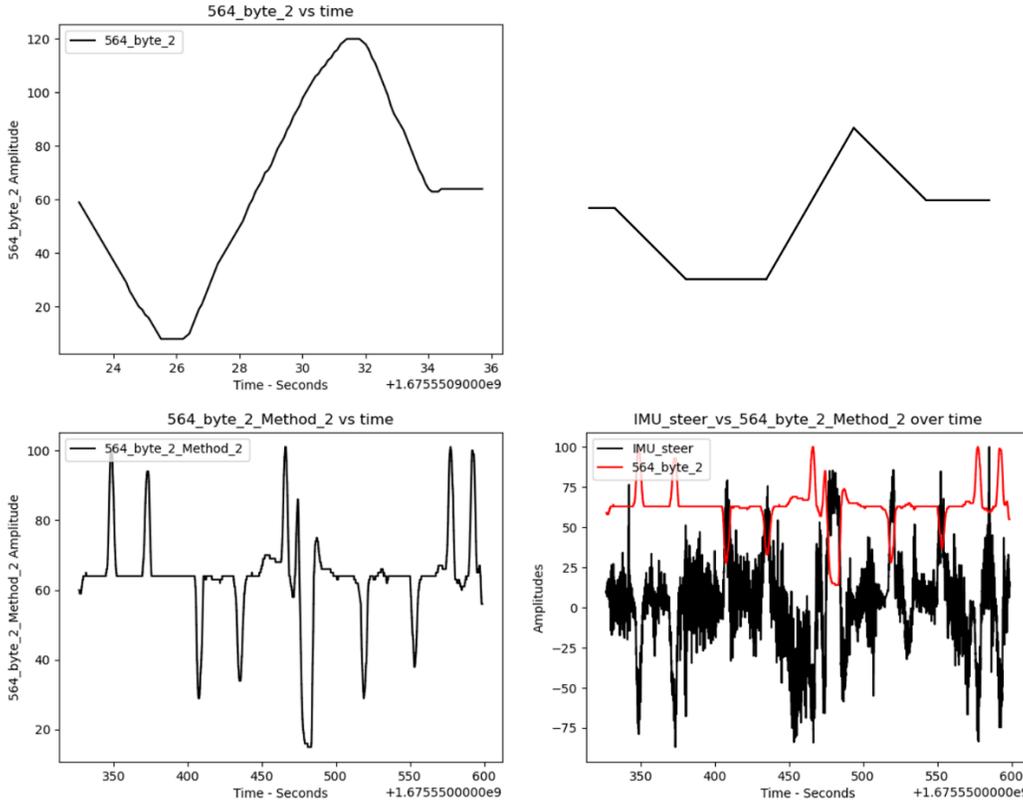

**Figure A4.** 2021 GMC Sierra 2500 steering wheel position verification: 564 byte 2 with GPS.

*A.2. 2022 Chevrolet Traverse Results*

This section presents data for the 2022 Chevrolet Traverse. Table A4 lists the best correlating CAN channels for the brake pedal. Figure A5 illustrates the results for CAN 190_msb_sixteen_bit_1 and Figure A6 illustrates the results for CAN channel 241_byte_1. For each figure, the top images show the CAN channel data during the accelerator pedal calibration recording. The bottom images show the CAN channel data during the vehicle trip recording.

**Table A4.** 2022 Chevrolet Traverse results brake pedal position with GPS.

| ID | Channel | Correlation | Range | Unique | StDev(*) | Smooth |
|---|---|---|---|---|---|---|
| 241 | lsb_sixteen_bit_0 | 0.827471914 | 34942 | 315 | 219 | 1 |
| 241 | msb_ten_bit_3 | 0.826576491 | 272 | 135 | 2 | 1 |
| 241 | lsb_ten_bit_1 | 0.826576491 | 136 | 135 | 1 | 1 |
| 241 | lsb_nine_bit_1 | 0.826576491 | 136 | 135 | 1 | 1 |
| 241 | msb_thirteen_bit_3 | 0.826576491 | 272 | 135 | 2 | 1 |
| 241 | lsb_fourteen_bit_1 | 0.826576491 | 136 | 135 | 1 | 1 |
| 241 | msb_sixteen_bit_3 | 0.826576491 | 272 | 135 | 2 | 1 |
| 241 | byte_1 | 0.826576491 | 136 | 135 | 1 | 1 |
| 241 | msb_fifteen_bit_3 | 0.826576491 | 272 | 135 | 2 | 1 |
| 241 | msb_fourteen_bit_3 | 0.826576491 | 272 | 135 | 2 | 1 |

| 241 | lsb_thirteen_bit_1 | 0.826576491 | 136 | 135 | 1 | 1 |
| 241 | msb_eleven_bit_3 | 0.826576491 | 272 | 135 | 2 | 1 |
| 241 | msb_nine_bit_0 | 0.826576491 | 136 | 135 | 1 | 1 |
| 241 | lsb_eleven_bit_1 | 0.826576491 | 136 | 135 | 1 | 1 |
| 241 | lsb_twelve_bit_1 | 0.826576491 | 136 | 135 | 1 | 1 |
| 241 | msb_twelve_bit_3 | 0.826576491 | 272 | 135 | 2 | 1 |
| 241 | msb_nine_bit_3 | 0.826576491 | 272 | 135 | 2 | 1 |
| 241 | msb_sixteen_bit_1 | 0.826575492 | 34816 | 189 | 260 | 1 |
| 190 | msb_sixteen_bit_1 | 0.817322609 | 34304 | 127 | 327 | 1 |
| 190 | lsb_sixteen_bit_0 | 0.817261267 | 34316 | 263 | 275 | 1 |

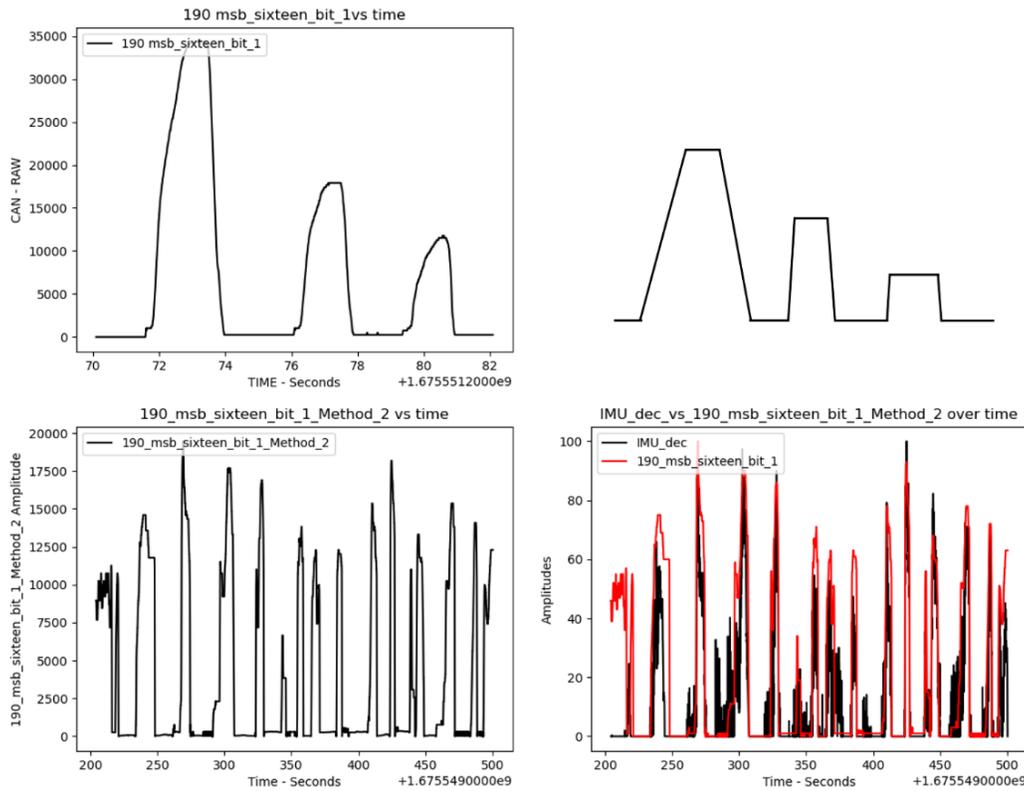

**Figure A5.** 2022 Chevrolet Traverse brake pedal position channel 190_msb_sixteen_bit_1 with GPS.

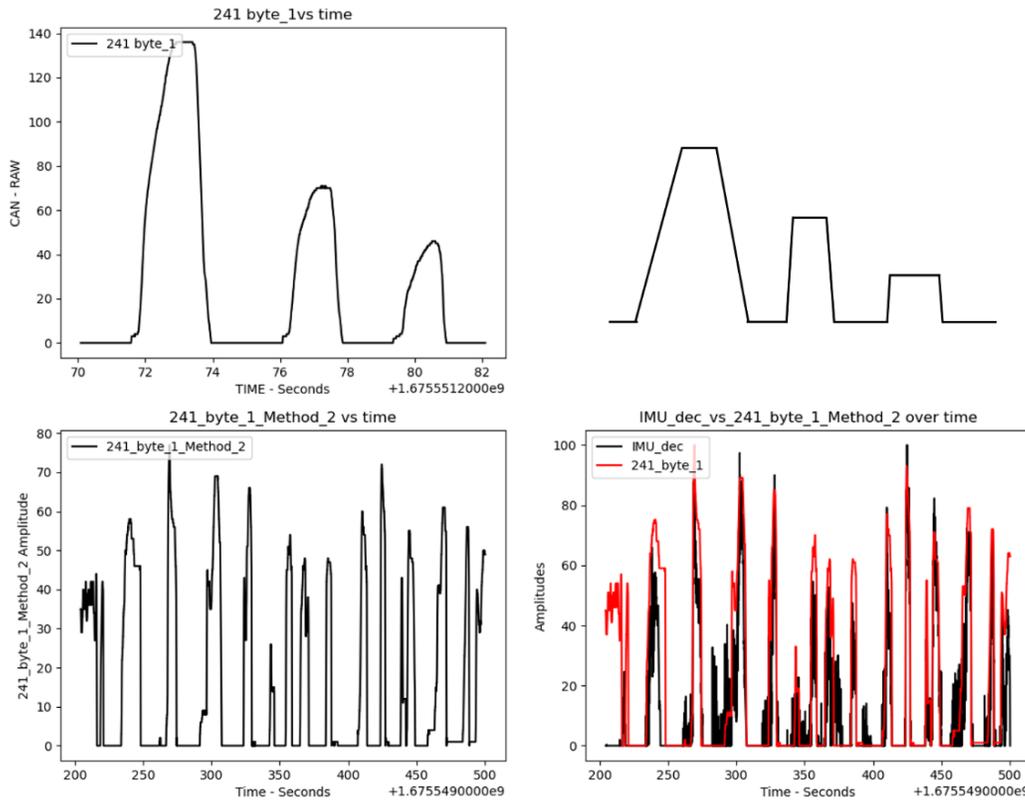
**Figure A6.** 2022 Chevrolet Traverse brake pedal position channel 241_byte_1 with GPS.

Table A5 lists the best corelating CAN channels for the steering wheel position found using the first method (without GPS data) for the 2022 Chevrolet Traverse. Figure A7 shows the results for CAN channel 564_byte_2. The top images show the CAN channel data during the accelerator pedal calibration recording. The bottom images show the CAN channel data during the vehicle trip recording.

**Table A5.** 2022 Chevrolet Traverse results steering wheel position without GPS.

| ID | Channel | Correlation | Range | Unique | StDev(*) | Smooth |
|---|---|---|---|---|---|---|
| 564 | lsb_sixteen_bit_1 | 0.620322776 | 32000 | 85 | 180 | 1 |
| 564 | msb_sixteen_bit_2 | 0.620307824 | 32000 | 85 | 180 | 1 |
| 564 | msb_fifteen_bit_2 | 0.620307824 | 32000 | 85 | 180 | 1 |
| 564 | byte_2 | 0.620307797 | 125 | 85 | 1 | 1 |
| 564 | lsb_nine_bit_2 | 0.620307797 | 125 | 85 | 1 | 1 |
| 564 | lsb_sixteen_bit_2 | 0.539768763 | 125 | 85 | 1 | 1 |

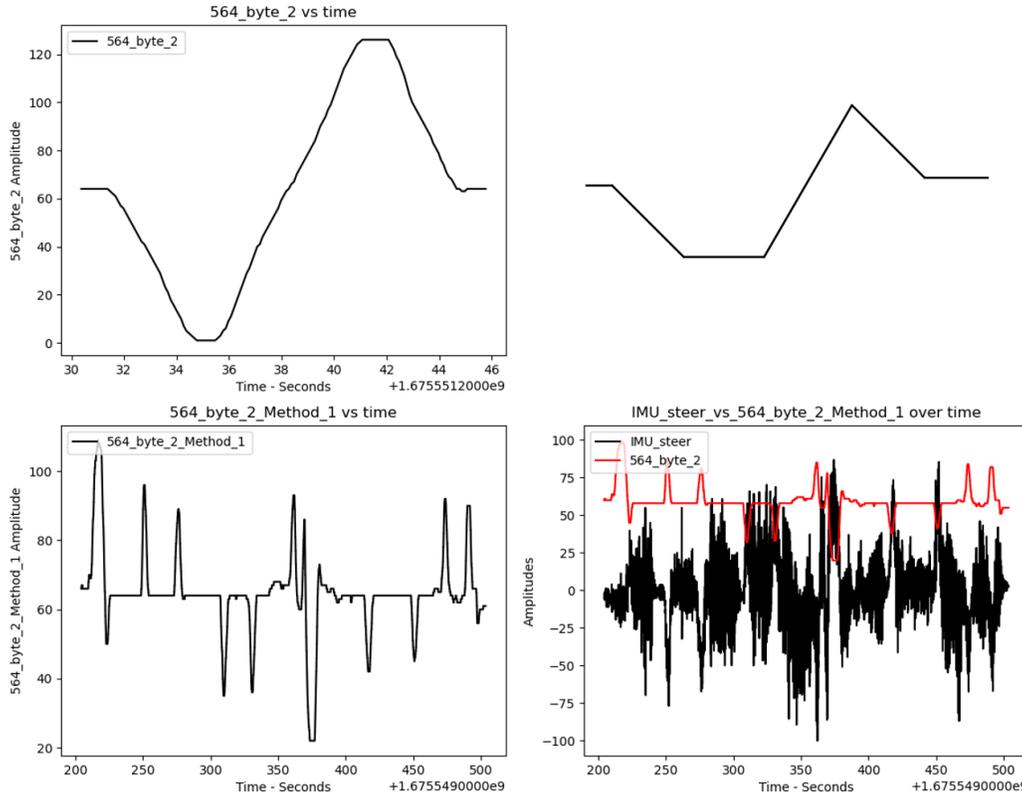

**Figure A7.** 2022 Chevrolet Traverse steering wheel position channel 564_byte_2 without GPS.

Table A6 lists the best correlating CAN channels for the steering wheel position found using the second method (with GPS) for the 2022 Chevrolet Traverse. Figure A8 illustrates the results for CAN channel 564_byte_2. The top images show the CAN channel data during the accelerator pedal calibration recording. The bottom images show the CAN channel data during the vehicle trip recording.

**Table A6.** 2022 Chevrolet Traverse results steering wheel position with GPS.

| ID | Channel | Correlation | Range | Unique | StDev(*) | Smooth |
|---|---|---|---|---|---|---|
| 564 | lsb_sixteen_bit_1 | 0.667576096 | 32000 | 85 | 180 | 1 |
| 564 | msb_sixteen_bit_2 | 0.667569894 | 32000 | 85 | 180 | 1 |
| 564 | msb_fifteen_bit_2 | 0.667569894 | 32000 | 85 | 180 | 1 |
| 564 | byte_2 | 0.667569855 | 125 | 85 | 1 | 1 |
| 564 | lsb_nine_bit_2 | 0.667569855 | 125 | 85 | 1 | 1 |
| 564 | lsb_sixteen_bit_2 | 0.56661827 | 125 | 85 | 1 | 1 |

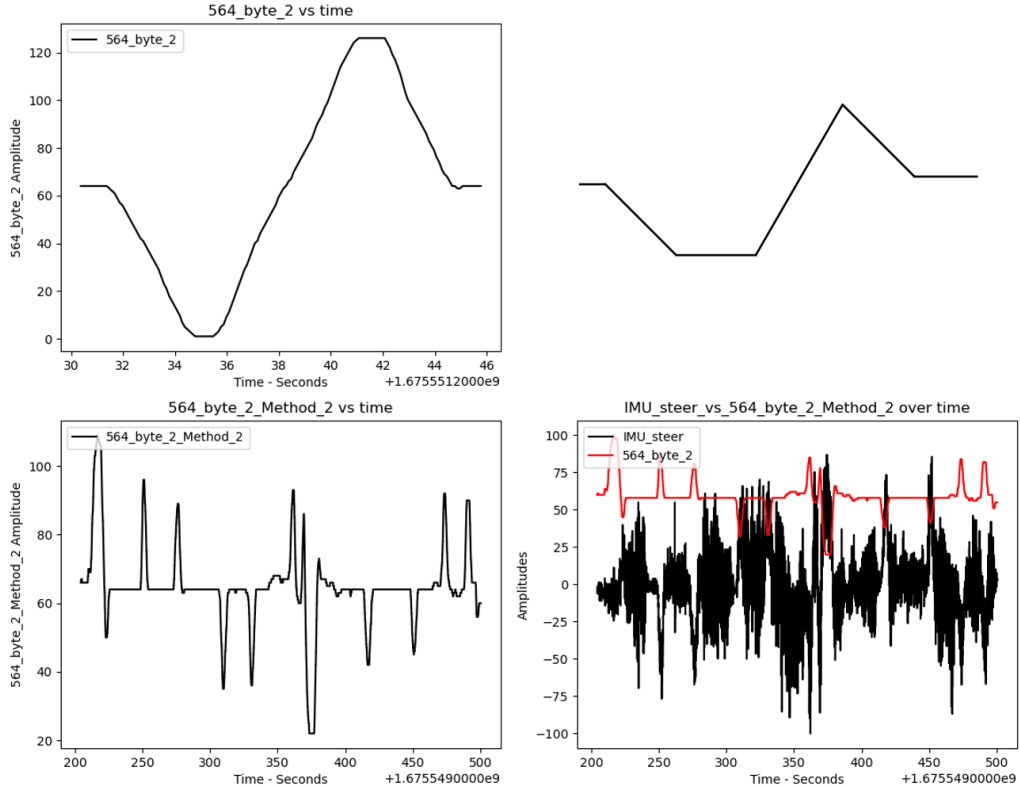

**Figure A8.** 2022 Chevrolet Traverse steering wheel position channel 564_byte_2 with GPS.

*A.3. 2009 Chevrolet Impala Results*

This section presents the data for the 2009 Chevrolet Impala. Table A7 lists the best correlating CAN channels for the brake pedal position. Figure A9 shows the results for CAN 241_lsb_sixteen_bit_0 and Figure A10 presents illustrates the results for CAN channel 241_msb_sixteen_bit_1. For each figure, the top images show the CAN channel data during the accelerator pedal calibration recording. The bottom images show the CAN channel data during the vehicle trip recording.

Table A7. Chevrolet Impala results brake pedal position with GPS.

| ID  | Channel           | Correlation | Range | Unique | StDev(*) | Smooth |
|-----|-------------------|-------------|-------|--------|----------|--------|
| 241 | lsb_sixteen_bit_0 | 0.757060037 | 33662 | 412    | 224      | 1      |
| 241 | msb_sixteen_bit_1 | 0.756210636 | 33536 | 278    | 265      | 1      |

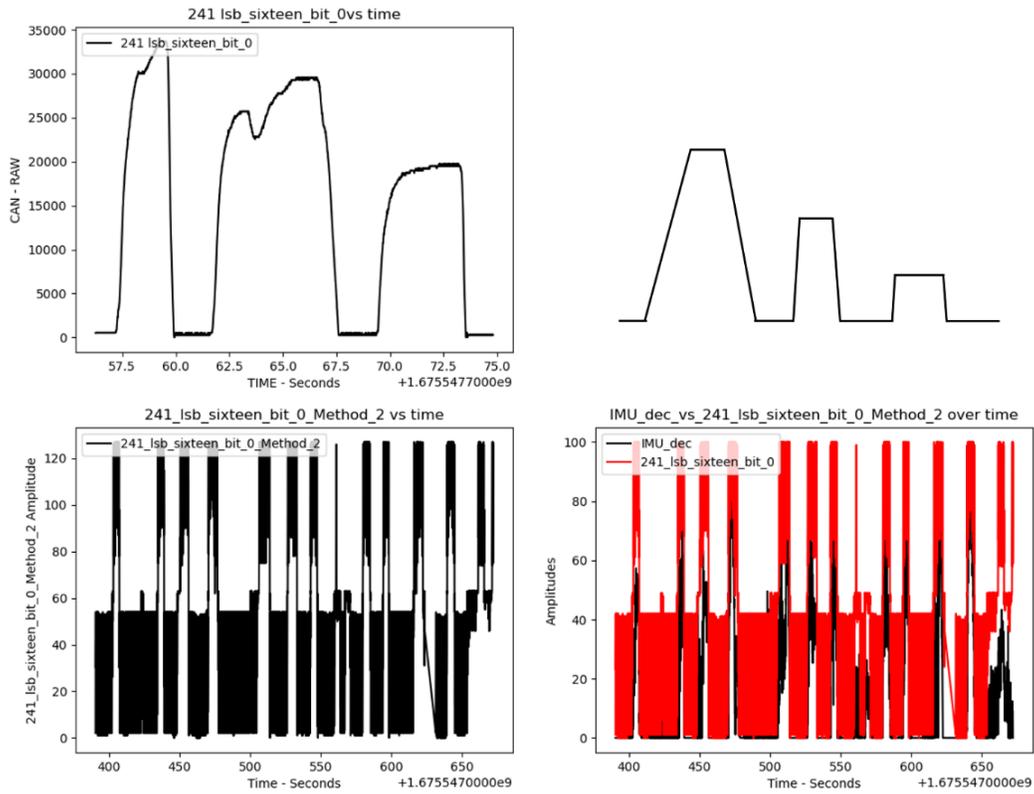

**Figure A9.** 2009 Chevrolet Impala brake pedal position channel 241_lsb_sixteen_bit_0 with GPS.

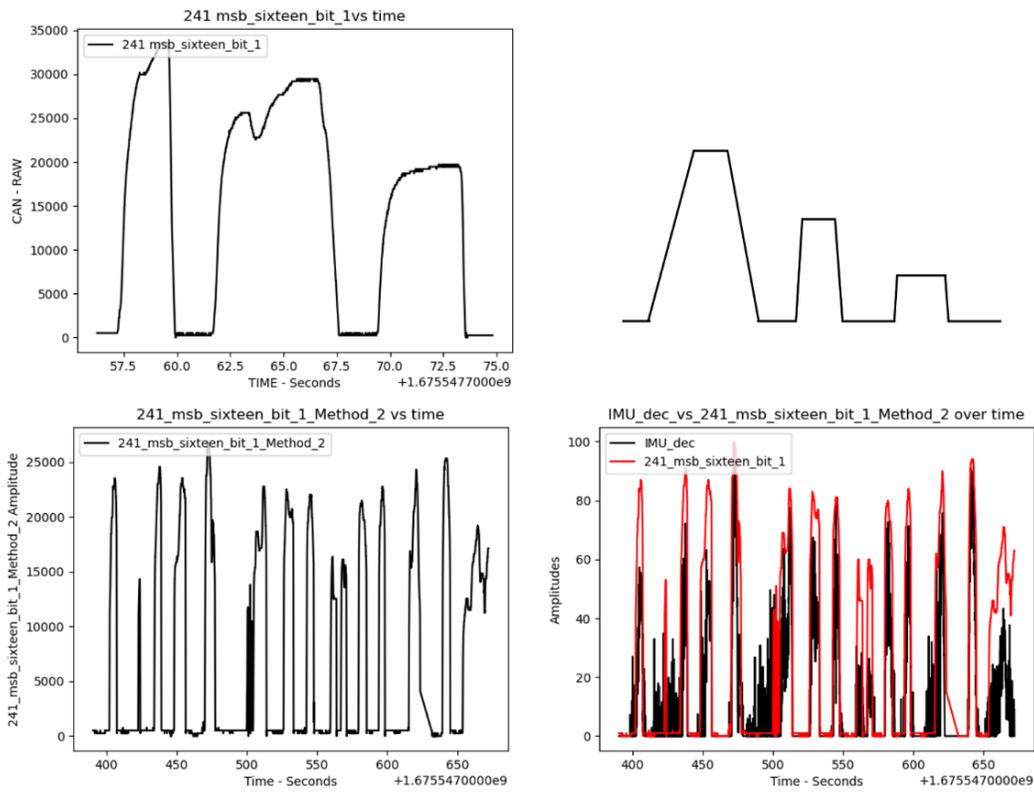

**Figure A10.** 2009 Chevrolet Impala brake pedal position channel 241_msb_sixteen_bit_1with GPS.

No steering wheel position CAN channels were identified for the 2009 Chevrolet Impala.

*A.4. 2006 Volvo XC90 Results*

This section presents the data for the 2009 Chevrolet Impala. Table A8 lists the best correlating CAN channels for the brake pedal position. Figure A11 illustrates the results for CAN 2244644_msb_sixteen_bit_2 and Figure A12 shows the results for CAN channel 2244644_msb_ten_bit_6. For each figure, the top images show the CAN channel data during the accelerator pedal calibration recording. The bottom images show the CAN channel data during the vehicle trip recording.

**Table A8.** 2006 Volvo XC90 results brake pedal position with GPS.

| ID | Channel | Correlation | Range | Unique | StDev(*) | Smooth |
|---|---|---|---|---|---|---|
| 2244644 | msb_eleven_bit_6 | 0.915151442 | 740 | 353 | 6 | 1 |
| 2244644 | msb_ten_bit_6 | 0.915151442 | 740 | 353 | 6 | 1 |
| 2244644 | msb_sixteen_bit_2 | 0.752932036 | 25088 | 99 | 164 | 1 |
| 2244644 | msb_fifteen_bit_2 | 0.752932036 | 25088 | 99 | 164 | 1 |
| 2244644 | lsb_sixteen_bit_1 | 0.752856582 | 25088 | 99 | 164 | 1 |

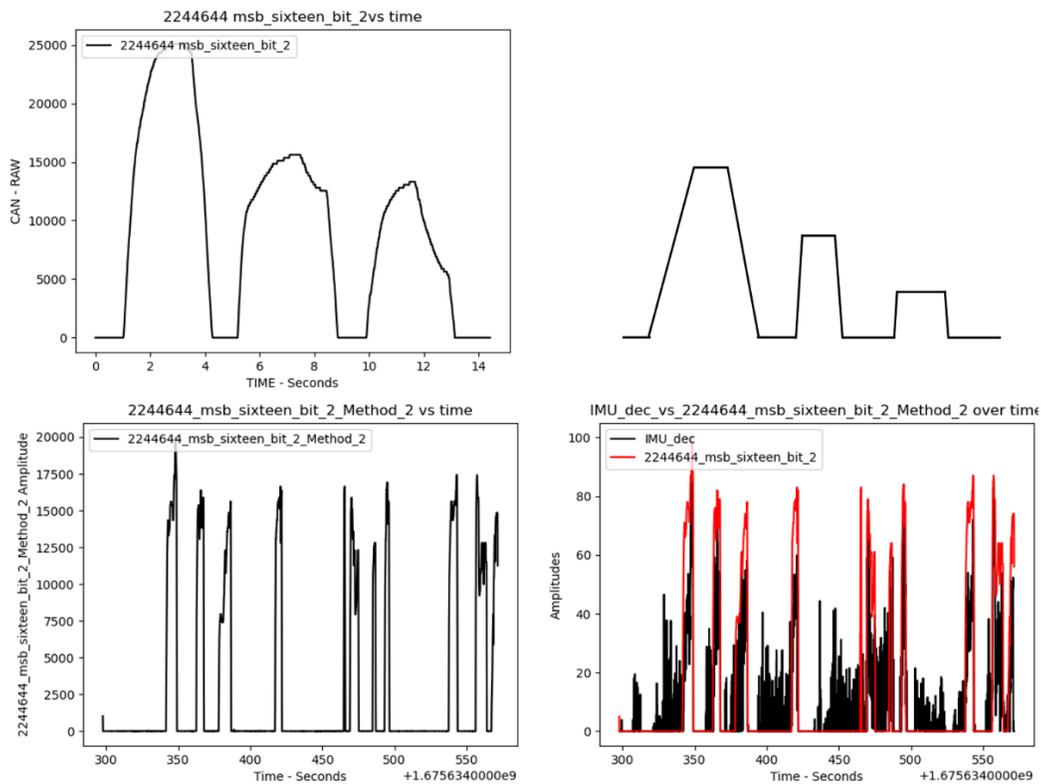

**Figure A11.** 2006 Volvo XC90 brake pedal position channel 2244644_msb_sixteen_bit_2 with GPS.

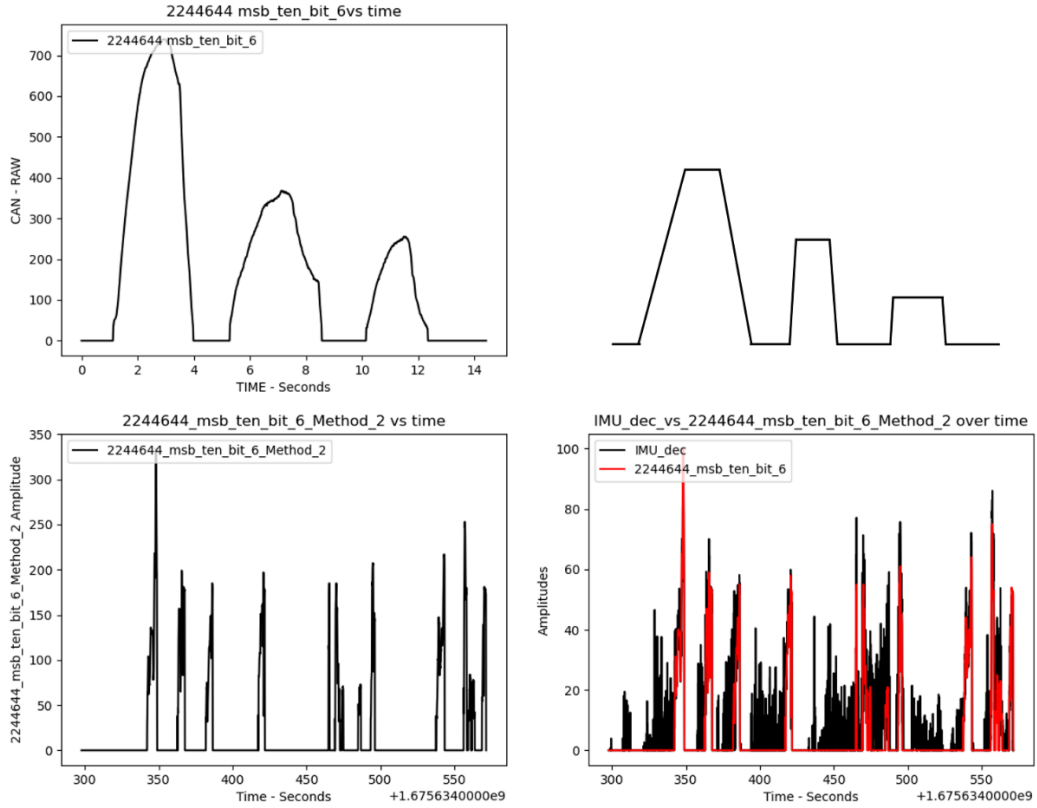

**Figure A12.** 2006 Volvo XC90 brake pedal position channel 2244644_msb_ten_bit_6 with GPS.

Table A9 lists the best correlating CAN channels for the steering wheel position found using the first method, (without GPS) for the 2006 Volvo XC90. Figure A13 illustrates the results for CAN channel 2244644_msb_ten_bit_6. The top images show the CAN channel data during the accelerator pedal calibration recording. The bottom images show the CAN channel data during the vehicle trip recording.

**Table A9.** 2006 Volvo XC90 results steering wheel position without GPS.

| ID | Channel | Correlation | Range | Unique | StDev(*) | Smooth |
|---|---|---|---|---|---|---|
| 283262976 | msb_fifteen_bit_2 | 0.775788374 | 1621 | 244 | 6 | 1 |
| 283262976 | msb_sixteen_bit_2 | 0.775788374 | 1621 | 244 | 6 | 1 |
| 283262976 | msb_thirteen_bit_2 | 0.775788374 | 1621 | 244 | 6 | 1 |
| 283262976 | msb_twelve_bit_2 | 0.775788374 | 1621 | 244 | 6 | 1 |
| 283262976 | msb_fourteen_bit_2 | 0.775788374 | 1621 | 244 | 6 | 1 |

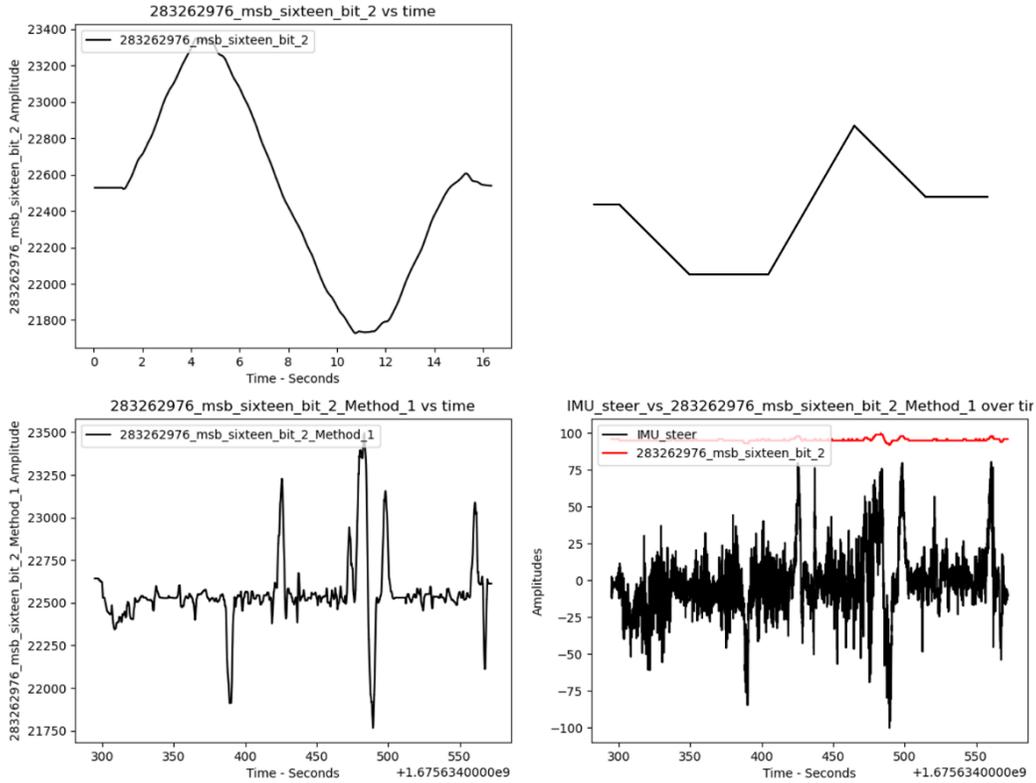

**Figure A13.** 2006 Volvo XC90 steering wheel position channel 2244644_msb_ten_bit_6 with GPS.

Table A10 lists the best correlating CAN channels for the steering wheel position found using the second method (with GPS) for the 2006 Volvo XC90. Figure A14 shows the results for CAN channel 28326976_msb_sixteen_bit_2. The top images show the CAN channel data during the accelerator pedal calibration recording. The bottom images show the CAN channel data during the vehicle trip recording.

**Table A10.** 2006 Volvo XC90 results steering wheel position with GPS.

| ID | Channel | Correlation | Range | Unique | StDev(*) | Smooth |
|---|---|---|---|---|---|---|
| 283262976 | msb_twelve_bit_2 | 0.814687756 | 1621 | 244 | 6 | 1 |
| 283262976 | msb_fifteen_bit_2 | 0.814687756 | 1621 | 244 | 6 | 1 |
| 283262976 | msb_sixteen_bit_2 | 0.814687756 | 1621 | 244 | 6 | 1 |
| 283262976 | msb_fourteen_bit_2 | 0.814687756 | 1621 | 244 | 6 | 1 |
| 283262976 | msb_thirteen_bit_2 | 0.814687756 | 1621 | 244 | 6 | 1 |

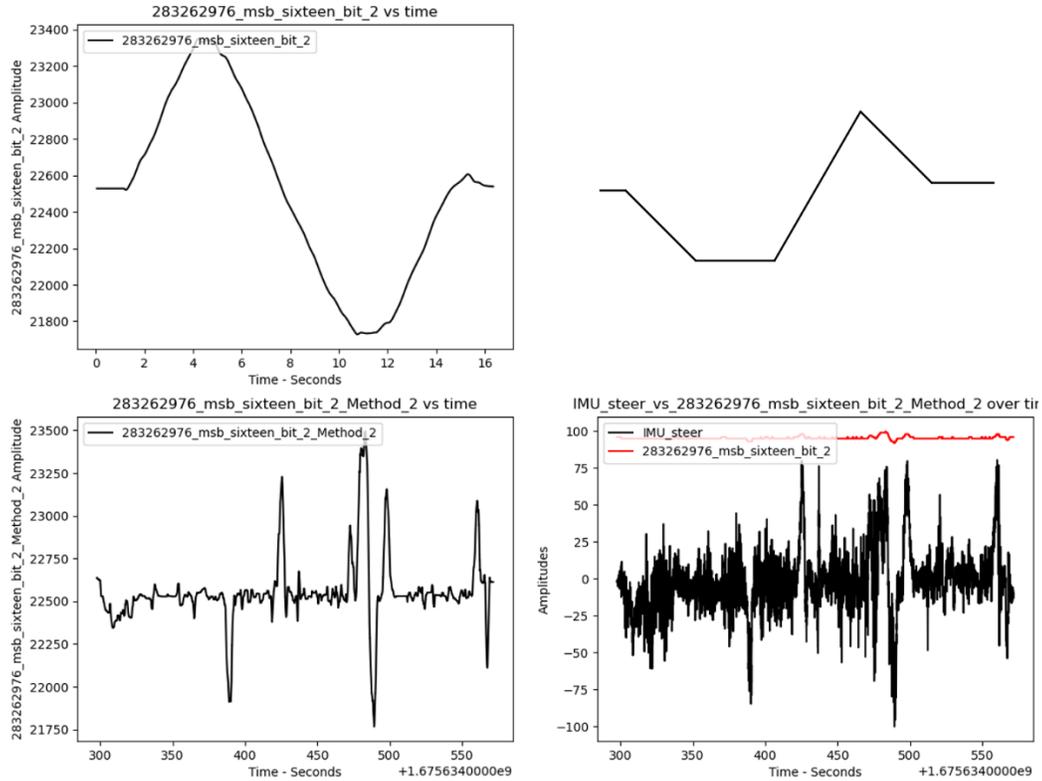

**Figure A14.** 2006 Volvo XC90 steering wheel position channel 28326976_msb_sixteen_bit_2 with GPS.

*A.5. 2016 Ford Fusion Results*

This section presents data for the 2016 Ford Fusion. Table A11 lists the best correlating CAN channels for the brake pedal position. Figure A15 illustrates the results for CAN 125_msb_ten_bit_3 and Figure A16 shows the results for CAN channel 125_msb_twelve_bit_3. For each figure, the top images show the CAN channel data during the accelerator pedal calibration recording. The bottom images show the CAN channel data during the vehicle trip recording.

**Table A11.** 2016 Ford Fusion results brake pedal position with GPS.

| ID | Channel | Correlation | Range | Unique | StDev(*) | Smooth |
|---|---|---|---|---|---|---|
| 125 | msb_ten_bit_3 | 0.95991078 | 645 | 223 | 12 | 2 |
| 125 | msb_eleven_bit_3 | 0.95991078 | 645 | 223 | 12 | 2 |
| 125 | msb_twelve_bit_3 | 0.95991078 | 645 | 223 | 12 | 2 |

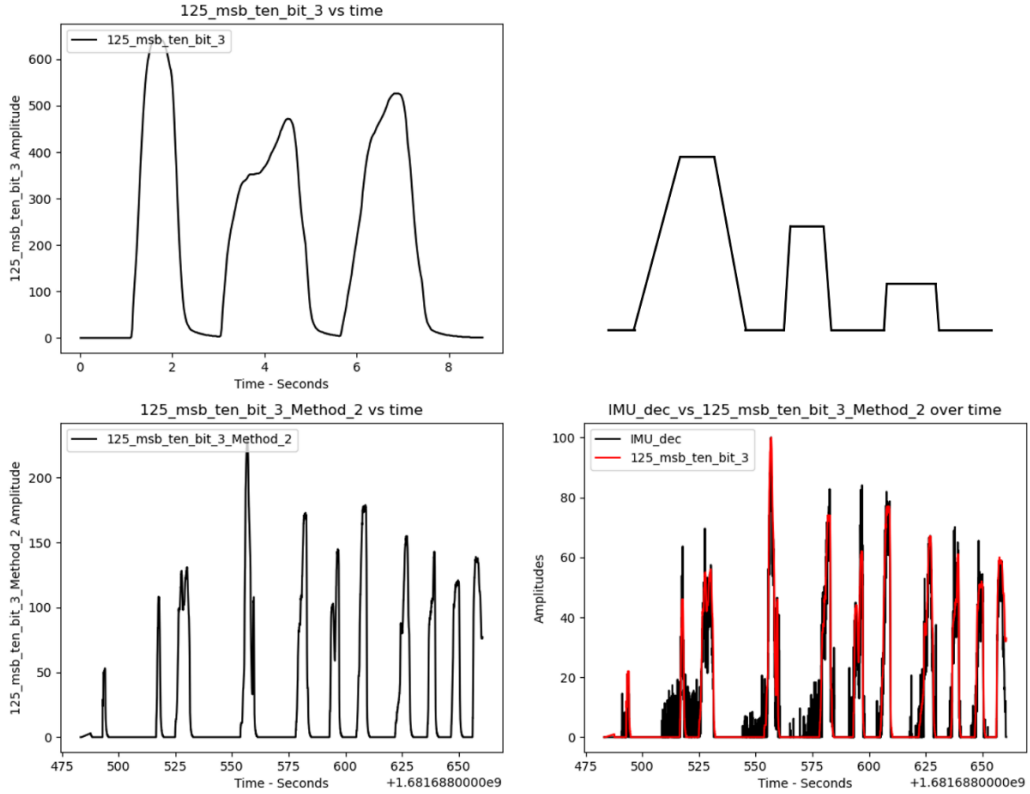

**Figure A15.** 2016 Ford Fusion brake pedal position channel 125_msb_ten_bit_3 with GPS.

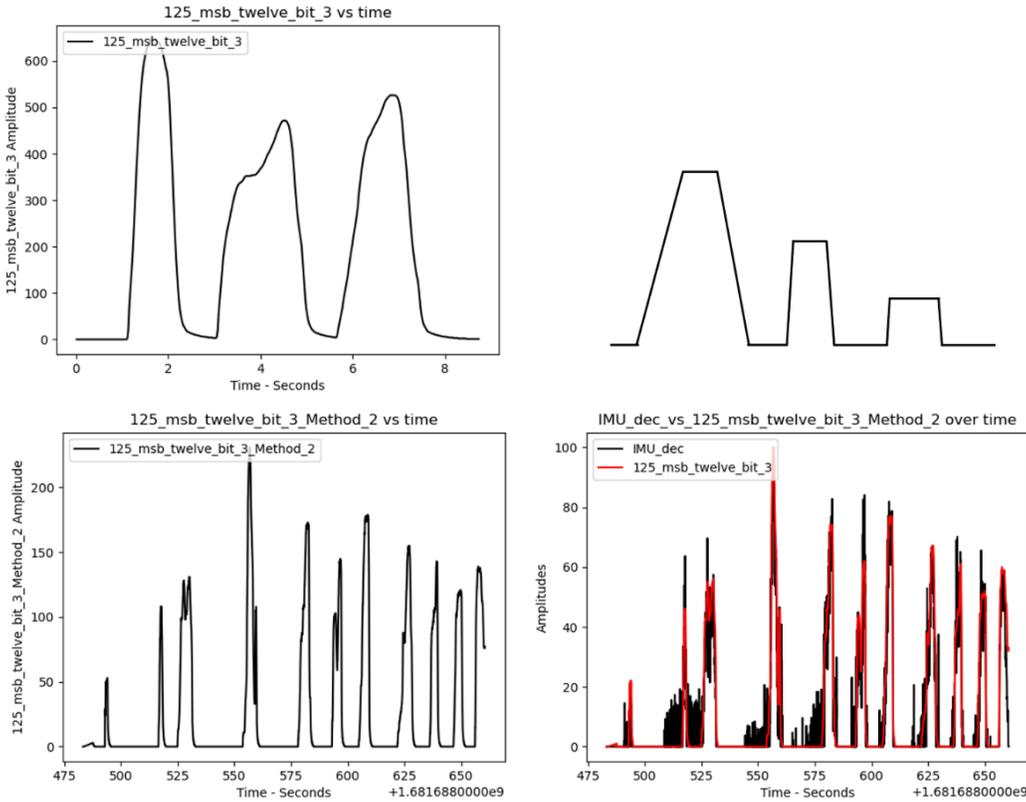

**Figure A16.** 2016 Ford Fusion brake pedal position channel 125_msb_twelve_bit_3 with GPS.

Table A12 lists the best correlating CAN channels for the steering wheel position found using the first method (without GPS) for the 2016 Ford Fusion. Figure A17 shows the results for CAN channel 118_msb_sixteen_bit_0. The top images show the CAN channel data during the accelerator pedal calibration recording. The bottom images show the CAN channel data during the vehicle trip recording.

**Table A12.** 2016 Ford Fusion results steering wheel position without GPS.

| ID | Channel | Correlation | Range | Unique | StDev(*) | Smooth |
|---|---|---|---|---|---|---|
| 118 | msb_sixteen_bit_0 | 0.685348995 | 9610 | 414 | 38 | 1 |
| 118 | msb_fifteen_bit_0 | 0.685348995 | 9610 | 414 | 38 | 1 |
| 133 | msb_sixteen_bit_0 | 0.684216386 | 9614 | 760 | 19 | 1 |

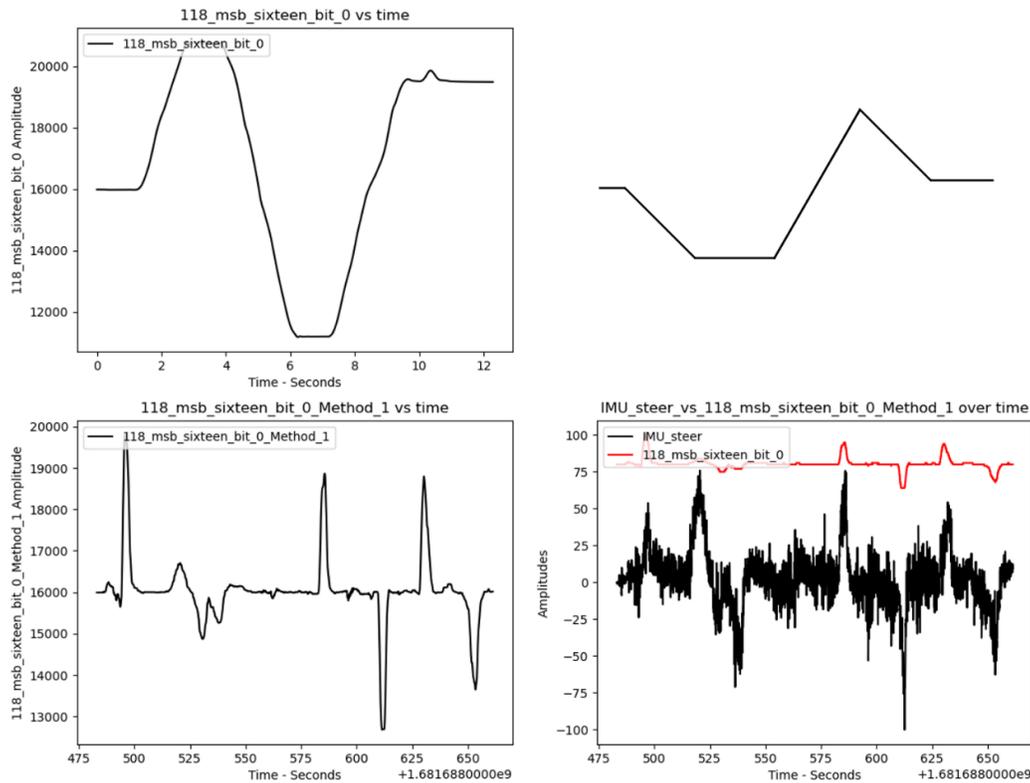

**Figure A17.** 2016 Ford Fusion steering wheel position channel 118_msb_sixteen_bit_0 without GPS.

Table A13 shows the best correlating CAN channels for the steering wheel position found using the second method (with GPS) for the 2016 Ford Fusion. Figure A18 shows the results for CAN channel 118_msb_sixteen_bit_0. The top images show the CAN channel data during the accelerator pedal calibration recording. The bottom images show the CAN channel data during the vehicle trip recording.

**Table A13.** 2016 Ford Fusion results steering wheel position with GPS.

| ID | Channel | Correlation | Range | Unique | StDev(*) | Smooth |
|---|---|---|---|---|---|---|
| 118 | msb_fifteen_bit_0 | 0.685702597 | 9610 | 414 | 38 | 1 |
| 118 | msb_sixteen_bit_0 | 0.685702597 | 9610 | 414 | 38 | 1 |
| 133 | msb_sixteen_bit_0 | 0.684574262 | 9614 | 760 | 19 | 1 |

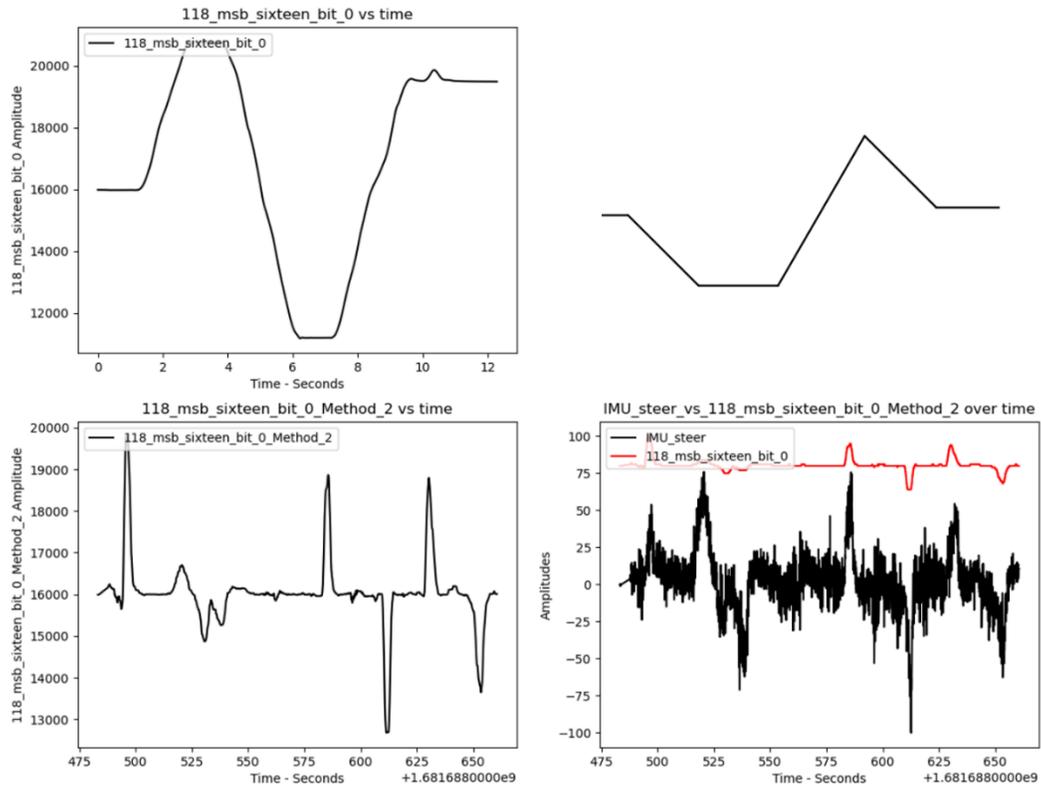

**Figure A18.** 2016 Ford Fusion steering wheel position channel 118_msb_sixteen_bit_0 with GPS.